\documentclass[conference]{IEEEtran}
\IEEEoverridecommandlockouts
\usepackage{cite}
\usepackage{amsmath,amssymb,amsfonts}
\usepackage{algorithmic}
\usepackage{graphicx}
\usepackage{textcomp}
\usepackage{xcolor}
\def\BibTeX{{\rm B\kern-.05em{\sc i\kern-.025em b}\kern-.08em
    T\kern-.1667em\lower.7ex\hbox{E}\kern-.125emX}}

\makeatletter
\newcommand{\linebreakand}{%
  \end{@IEEEauthorhalign}
  \hfill\mbox{}\par
  \mbox{}\hfill\begin{@IEEEauthorhalign}
}
\makeatother

\title{FinXABSA: Explainable Finance \\through Aspect-Based Sentiment Analysis}
\author{
\IEEEauthorblockN{\hspace{1cm}Keane Ong}
\IEEEauthorblockA{\hspace{1cm}\textit{Asian Institute of Digital Finance} \\
\textit{\hspace{1cm}National University of Singapore}\\
\hspace{1cm}3 Research Link, Singapore 117602 \\
\hspace{1cm}keane.ongweiyang@u.nus.edu}
\and
\IEEEauthorblockN{\hspace{1.5cm}Wihan van der Heever}
\IEEEauthorblockA{\hspace{1.5cm}\textit{School of Computer Science and Engineering} \\
\textit{\hspace{1.5cm}Nanyang Technological University}\\
\hspace{1.5cm}50 Nanyang Ave, Singapore 639798 \\
\hspace{1.5cm}wihan001@e.ntu.edu.sg}
\hfill\linebreakand
\IEEEauthorblockN{Ranjan Satapathy}
\IEEEauthorblockA{\textit{Institute of High Performance Computing (IHPC)}\\
\textit{Agency for Science, Technology and Research (A\textasteriskcentered STAR)}\\
Fusionopolis Way, \#16-16 Connexis, Singapore 138632\\
satapathy\_ranjan@ihpc.a-star.edu.sg}
\and
\IEEEauthorblockN{Erik Cambria}
\IEEEauthorblockA{\textit{School of Computer Science and Engineering} \\
\textit{Nanyang Technological University}\\
50 Nanyang Ave, Singapore 639798 \\
cambria@ntu.edu.sg}
\hfill\linebreakand
\IEEEauthorblockN{Gianmarco Mengaldo}
\IEEEauthorblockA{\textit{College of Design \& Engineering} \\
\textit{National University of Singapore}\\
9 Engineering Drive 1, Singapore 117575\\
mpegim@nus.edu.sg}}

\begin{document} 

\maketitle

\begin{abstract}
This paper presents a novel approach for explainability in financial analysis by deriving financially-explainable statistical relationships through aspect-based sentiment analysis, Pearson correlation, Granger causality \& uncertainty coefficient. The proposed methodology involves constructing an aspect list from financial literature and applying aspect-based sentiment analysis on social media text to compute sentiment scores for each aspect. Pearson correlation is then applied to uncover financially explainable relationships between aspect sentiment scores and stock prices. Findings for derived relationships are made robust by applying Granger causality to determine the forecasting ability of each aspect sentiment score for stock prices. Finally, an added layer of interpretability is added by evaluating uncertainty coefficient scores between aspect sentiment scores and stock prices. This allows us to determine the aspects whose sentiment scores are most statistically significant for stock prices. Relative to other methods, our approach provides a more informative and accurate understanding of the relationship between sentiment analysis and stock prices. Specifically, this methodology enables an interpretation of the statistical relationship between aspect-based sentiment scores and stock prices, which offers explainability to AI-driven financial decision-making. 
\end{abstract}

\section{Introduction}
Given the increasing use of AI in finance, explainability has become crucial for transparency, trust, and accountability in financial decisions. Interpretable AI aids in spotting errors and biases, bolstering investor confidence and AI credibility ~\cite{rudin2019stop}. 
This paper presents an explainable AI method for finance, merging aspect-based sentiment analysis (ABSA) with statistical techniques. After extracting financial aspects, we use the ABSA model from~\cite{liang2022aspect} to compute sentiment scores and analyze their correlation with specific stocks. 

Our study addresses three core questions: the potential for significant explainable correlations through this framework, the application of Granger causality for robust correlations, and the use of the uncertainty coefficient to interpret financial aspect significance. 
Our results highlight key financial aspects impacting stock prices and establish intelligible and robust statistical relationships. The Pearson correlation with ABSA reveals an explainable link between financial aspect sentiment and stock prices, reinforced by the Granger causality test. The entropy-based uncertainty coefficient further pinpoints the most influential aspect sentiment scores for stock prices.

The paper is organized as follows: Section~\ref{section: lit_review} investigates research in financial sentiment analysis; 
Section~\ref{section: data} focuses on data collection; Section~\ref{section: sgcn} delineates the model adopted; Section \ref{section: sentiment_scores} explains our calculation of aspect sentiment scores; Sections~\ref{section: stat_methods_pc},~\ref{section: stat_methods_gc} \&~\ref{section: stats_methods_uc} briefly describe statistical methods employed; Section~\ref{section: results} highlights the main results; Section~\ref{section: discussion} outlines main discussion points; finally, Section~\ref{section: conclusion} discusses limitations and future research directions.  

\section{Related Work}~\label{section: lit_review}
Sentiment analysis is a natural language processing (NLP) technique leveraging computational methods to determine the polarity or emotional tone expressed in a piece of text~\cite{onesta}. Different AI techniques have been leveraged to improve both accuracy and interpretability of sentiment analysis algorithms, including symbolic AI~\cite{camcomshort,xincog}, subsymbolic AI~\cite{shortcammed,camsta}, and neurosymbolic AI~\cite{valcon,camsev}. 
Recent work on ABSA include~\cite{mao2021bridging}, which combines multitask learning with ABSA, and~\cite{he2022meta}, which adopts a meta-weighting strategy. 

\begin{figure}[h]
   ~\centering
   ~\includegraphics[width=\linewidth]{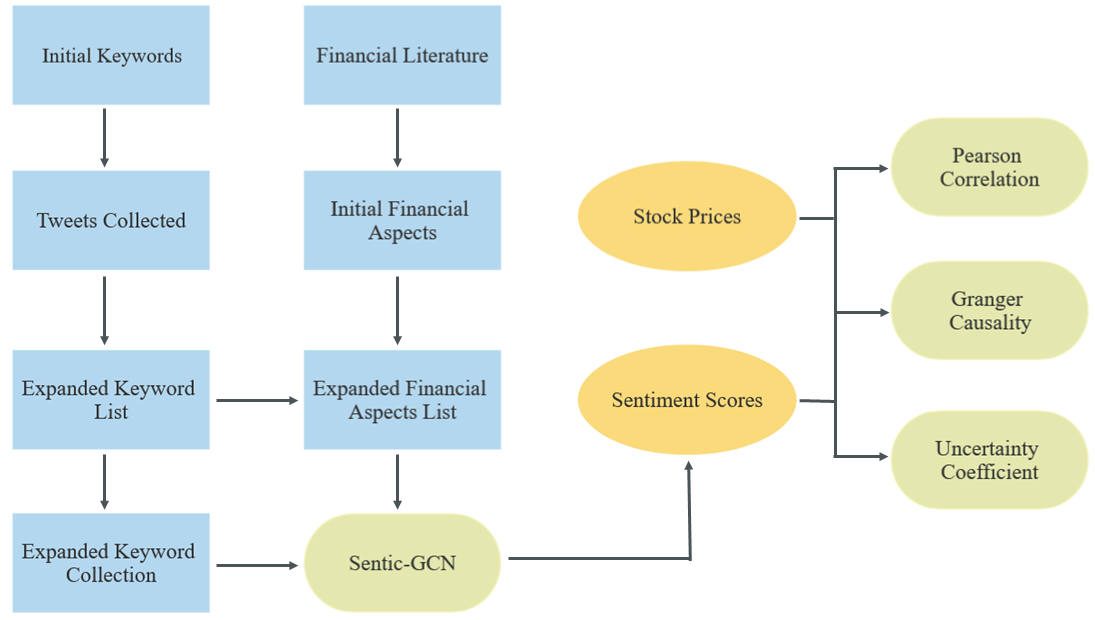}
   ~\caption{The architecture of the proposed method for XFSA.}
   ~\label{fig:finxai_fig}
\end{figure}

Finance-specific ABSA techniques used CNNs and transfer learning~\cite{jangid2018aspect,yang2018financial}. Financial Sentiment Analysis (FSA) methods are more comprehensively covered in~\cite{man2019financial}, and pertinent for investment. \cite{ma2023multi} emphasizes market news sentiment and company metrics for stock price forecasting, whereas~\cite{wang2023learning} leverages technical indicators and social media sentiment. 
\cite{picasso2019technical} and~\cite{wang2023learning} highlight FSA's utility for uncovering market trends.
FSA and ABSA have progressed, but explainable sentiment analysis in finance (XFSA) is still emerging. Recent developments pave the way for improved explainability which can enhance FSA's reliability in financial decisions. \cite{gite2021explainable} combined sentiment and technical analysis for clear stock predictions. \cite{luo2018beyond} introduced an interpretable neural net for FSA with a query-driven attention mechanism. \cite{loginova2021forecasting} emphasized better interpretability using ABSA for Bitcoin text forecasting.

Separately, statistical methods have proven useful for uncovering the dependence between financial variables. Pearson correlation can gauge sentiment-stock price links, as demonstrated by~\cite{ruan2018using}. Granger causality, discussed in~\cite{lee2014granger}, highlighted bi-directional causality between sentiment and the Chinese stock market~\cite{chu2016nonlinear}. Additionally, Pearson correlation and Granger causality have been jointly utilized in sentiment-stock price studies~\cite{hamraoui2022impact}. Uncertainty coefficients, highlighted by~\cite{kim2021information} and~\cite{birru2022sentiment}, reveal sentiment's influence on stock returns during high information uncertainty.


\section{Data Collection}~\label{section: data}
\subsection{Gathering Keywords via `\textit{Keyword Hopping}'}\label{subsect:keywordhopping}
Using a `keyword hopping' framework, we began with the keywords \textit{`nasdaq stock market'} to gather tweets from Q4 2022, leveraging NASDAQ's prominence with over 3,300 company listings. From approximately 11k tweets, we identified high-frequency keywords (above 100)~\cite{satapathy2017subjectivity}, counting each word once per tweet. After filtering out overly specific or non-financial terms like \textit{tesla} or \textit{cnbc}, we added relevant financial keywords such as \textit{`sharemarket', `stockstobuy'}~\cite{dhegra}. The comprehensive keyword list is detailed in the next section. 

\subsection{Twitter API}
Tweets were collected via the Twitter API v2 with academic access from Q4 2022. Due to the sheer number of tweets from the Twitter full archive search, we collected the tweets from only the turn of the hour for every hour each day. We utilised the following keywords:
\textit{stock market, Nasdaq, inflation, investors, friday sharemarket, monday sharemarket, china stock, china market, china economy, recession, Tuesday sharemarket, stock fall, thursday sharemarket, stock market, market rally, wednesday sharemarket, finance, economy, market closes, stock closes, financial market, sharemarket, stockstobuy, sharemarket drops, pandemic stock}. In our keyword list, commas mean `OR' and spaces mean `AND' for Twitter API queries. We excluded retweets and limited tweets to English, collecting about 120k tweets for sentiment analysis.

\subsection{Stock Prices}
We collected closing stock prices for six companies from Q4 2022, sourced from Yahoo Finance. Focusing on the rising significance of sustainable finance, we analyzed stocks from the sustainable energy sector, contrasting them with traditional energy. Traditional energy stocks are \textit{British Petroleum, Exxon, Shell}, and sustainable ones are \textit{NextEra, Clearway, Brookfield Renewable}. Each selected stock holds a significant market share.
Stock prices of companies reflect their financial health and are watched by investors and analysts. Analyzing this data reveals market trends and company performance. Fluctuations offer insights into financial stability, growth, and market position, aiding investors and experts in decision-making. In essence, the data on closing stock prices is vital for gauging the market and the companies' standings.

\subsection{Collecting financial aspects for Sentic GCN}
\begin{table}[b]
\caption{Top $20$ financial aspects}
    \centering
    \begin{tabular}{|c|c|c|c|}
    \hline
    Economic & Stock Market & Financial Institution & Corporate \\
    \hline 
        \textit{inflation} & \textit{investors} & \textit{finance} & \textit{report} \\
        \textit{economy} & \textit{market} & \textit{financial} & \textit{sales} \\
        \textit{recession} & \textit{stock} & \textit{rate} & \textit{cost} \\
        \textit{china} & \textit{trading} & \textit{interest} & \textit{tax} \\
         & \textit{price} & \textit{bank} & \\
         & \textit{stockmarket} & & \\
         & \textit{bitcoin} & & \\
    \hline    
    \end{tabular}
    \label{tab: top20aspects}
\end{table}

As explained later, a list of aspects (attributes or components of a sentence) is necessary for our task of ABSA. When working in the context of FSA, these aspects comprise lists of words used daily in the financial world, for example, ``\textit{share}", ``\textit{profit}" or ``\textit{risk}". An extensive list of aspects requires a large compilation of text data. Therefore, we exploit previous research in the FSA domain and draw upon the groundwork of~\cite{salunkhe2019aspect},~\cite{el2016learning} and~\cite{chen2017comparative}. The justification for making use of these existing aspect lists is that of trusted statistical methods to generate these words, such as Non-negative Matrix Factorisation, Latent Dirichlet Allocation, and Principal Component Analysis, and annotations by experts in the field, with~\cite{salunkhe2019aspect} sharing a similar goal to this paper. 

Additionally, as these words contain significant financial meaning, utilising them for ABSA enhances explainability. While keeping the words derived from the aforementioned financial literature the overwhelming bulk of our aspect list, we add another 24 financially important keywords from `Keyword Hopping' (\ref{subsect:keywordhopping}). Altogether, we assemble a list of 131 aspects to facilitate our analysis. Out of these, we focus on the 20 aspects that occurred most frequently in the text data (Table~\ref{tab: top20aspects}).

\section{Sentic GCN}~\label{section: sgcn}
\begin{figure}[b!]
   ~\centering
   ~\includegraphics[width=0.95\linewidth]{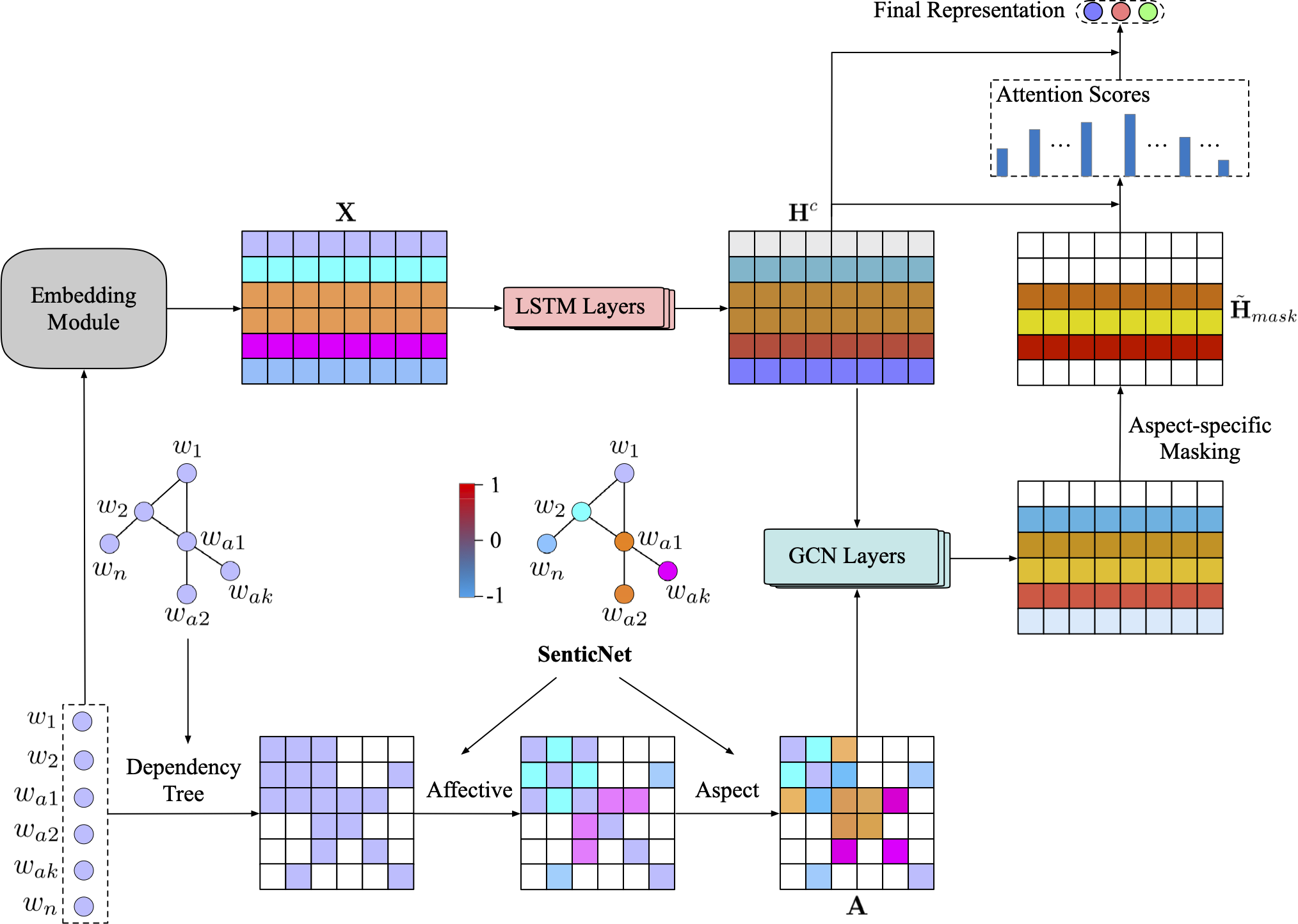}
   ~\caption{Sentic GCN architecture~\cite{liang2022aspect}}
   ~\label{fig:sgcn_fig}
\end{figure}

Sentic GCN~\cite{liang2022aspect} consists of two components, learning contextual representations and leveraging graph information. The first component is accomplished through LSTM layers, which derives latent contextual representations from the embedding matrix of each input sentence, while the second component entails utilising Graph Convolutional Network layers. These layers can express the potential sentiment dependencies of the contextual words by taking as input the hidden contextual representations, together with the matching affective enhanced graph. 

Thereafter, the model merges the representations of these two elements in order to deduce, with respect to a particular aspect, the most substantial dependencies. This improves upon majority of graph-based models which only considered the syntactical information contained within a sentence. Sentic GCN prioritises words with strong aspect-related sentiment by capitalising on the contextual sentiment dependencies concerning the specific aspect. This is done since the feature of aspect-related sentiment is crucial in ABSA tasks and as such the model refines the sentence's graph structure in an explicable manner. 
The entire process of the Sentic GCN model is illustrated in Fig.~\ref{fig:sgcn_fig}, where the final representation is the polarity of the different aspects of an input sentence. The depth of GCN layers is 2, L2 regularization coefficient $\lambda$ is 0.00001, Adam learning rate is 0.001, and hidden state vectors have 300 dimensionality.


\section{Sentiment Scores}~\label{section: sentiment_scores}
To compute the sentiment scores, Sentic GCN is employed on the collected tweets to label the sentiment of the collected financial aspects according to the polarities positive, neutral, and negative. We refer to absolute aspect sentiment scores as $x_{fp}$ \& $x_{fn}$. $x_{fp}$ is the number of times an aspect is labelled positive for each day, while $x_{fn}$ is the number of times an aspect is labelled negative for each day. $x_{fp}$ is referred to as positive absolute aspect sentiment score while $x_{fn}$ is referred to as negative absolute aspect sentiment score.

On the other hand, $x_{fs}$ is the sum of positive, negative and neutral labels corresponding to an aspect for each day. We refer to normalised aspect sentiment scores as $x_{nfp}$ \& $x_{nfn}$. $x_{nfp}$ is computed by dividing $x_{fp}$ with ${x_{fs}}$, while $x_{nfn}$ is computed by dividing ${x_{fn}}$ with ${x_{fs}}$. $x_{nfp}$ is referred to as positive normalised aspect sentiment score while $x_{nfn}$ is referred to as negative normalised aspect sentiment score.

In our paper, $\boldsymbol{x}$ can represent the different absolute aspect sentiment scores or normalised aspect sentiment scores (i.e. $\boldsymbol{x}=\{x_{fp},x_{fn}, x_{nfp}, x_{nfn}\}$). We also lag sentiment scores 1 day before stock prices before determining Pearson correlation, Granger causality \& uncertainty coefficient. The motivation for this follows from~\cite{smailovic2013predictive}, where lagging sentiment metrics behind stockprice has proven effective for reflecting price movements. Moreover we conduct statistical analyses only for the trading days of 2022 Q4.

\section{Pearson correlation}~\label{section: stat_methods_pc}
We will now explain the notation for stock prices. We represent the companies in our analysis by their stock symbol (i.e. Shell (SHEL), British Petroleum (BP), Exxon (XOM), Brookfield Renewable (BEPC), Clearway (CWEN), Nextera (NEE)). As such, our notation for daily stock closing price will contain the stock symbol. For example, ${y_{p,BP}}$ refers to the stock closing price of British Petroleum for the day, while ${y_{p,BEPC}}$ refers to the closing price of Brookfield Renewable for the day. $\boldsymbol{y_{p}}$ can represent the different stock closing prices (i.e. $\boldsymbol{y_{p}} = \{y_{p,SHEL}, y_{p,BP}, y_{p,XOM}, y_{p,BEPC}, y_{p,CWEN}, y_{p,NEE}\} $).

The Pearson correlation test is conducted to obtain the coefficient $r$, which measures the strength of linear relationship between two continuous variables. For our paper, we are conducting the correlation between sentiment scores lagged 1 day before stockpries. From equation (1), ${n}$ refers to the total number of \{$\boldsymbol{x,y_{p}}$\} pairs which is equivalent to the total trading days - 1 (due to the lag) of 2022 Q4. 

\begin{equation}
{r} =~\frac{\sum_{i=1}^n (\boldsymbol{x_i} -~\boldsymbol{\bar{x}})(\boldsymbol{y_{p,i}} -~\boldsymbol{\bar{y_{p}}})}{\sqrt{\sum_{i=1}^n (\boldsymbol{x_i} -~\boldsymbol{\bar{x}})^2\sum_{i=1}^n (\boldsymbol{y_{p,i}} -~\boldsymbol{\bar{y_{p}}})^2}}
\end{equation}

\vspace{0.5cm}

\section{Granger causality}~\label{section: stat_methods_gc}
The Granger causality test~\cite{granger1969investigating} highlights whether previous values of one variable encompasses data that helps predict another variable. In this paper, we will apply this test to uncover not only the forecasting ability of respective aspect sentiment scores for stock price prediction, but the interdependent relationship between these two variables. 
Particularly, we determine if the various aspect sentiment scores Granger cause the different stock prices. Our implementation of the Granger causality test is akin to the bivariate linear Granger causality test described in~\cite{hiemstra1994testing}. It is briefly explained below:

\begin{equation}
X_{t} = A(L) X_{t}+ B(L)Y_{p,t}+E_{X,t}
\end{equation}

\begin{equation}
Y_{p,t} = C(L) X_{t}+ D(L)Y_{p,t}+E_{Y_{p,t}}, \quad t=1,2, \ldots,
\end{equation}

${X_{t}}$ \& ${Y_{p,t}}$ can be the the time series of different sentiment scores, ${\boldsymbol{x}}$ \& different stock prices, $\boldsymbol{{y_{p}}}$ respectively. ${A(L)}$, ${B(L)}$, ${C(L)}$ \& ${D(L)}$ are the one-sided lag polynomials of \textit{a, b, c, d} respectively, where L is the lag operator. ${E_{X,t}}$ \& ${E_{Y_{p},t}}$ are the errors of regression. Granger causality is specifically tested via the F-test of exclusion restrictions, at the 0.05 significance level. $X_{t}$ Granger causes $Y_{p,t}$ if the constituents in $C(L)$ (that is, ${C_{i}}$ (i = 1, \ldots, c)) are jointly significantly different from zero. Vice-versa, $Y_{p,t}$ Granger causes $X_t$ if the constituents in $B(L)$, ${B_{i}}$ (i = 1, \ldots, b)) are jointly significantly different from zero. Although we have shown the bivariate implementation of the Granger causality test, in our paper, we only highlight Granger Causality in the direction of ${X_t}$ Granger causes ${Y_{p,t}}$. Particularly, we show the results for ${X_{t}}$ lagged one day behind $Y_{p,t}$. This is the most relevant to our analyses.

\section{Uncertainty coefficient}~\label{section: stats_methods_uc}

To supplement Granger causality and Pearson correlation, which primarily detect linear relationships, the uncertainty coefficient (also known as Theil's U or entropy coefficient) examines the statistical link between sentiment scores and stock prices without assuming linearity. This is crucial given the stock market's potential non-linear behavior~\cite{abhyankar1997uncovering}. The coefficient quantifies the reduction in entropy in stock prices, $\boldsymbol{y_{p}}$, when aspect sentiment scores, $\boldsymbol{x}$, are known. Essentially, it gauges the information about a stock price provided by an aspect sentiment score, denoted as ${U(\boldsymbol{y_{p}\mid x})}$. We calculate this using equations from Henri Theil~\cite{abramowitz1970handbook}, detailed in (4), (5), and (6).

\begin{equation}
{U(\boldsymbol{y_{p}\mid x})} = \frac{H(\boldsymbol{y_{p}}) - H(\boldsymbol{y_{p} \mid x})}{H(\boldsymbol{y_{p}})}
\end{equation}

${H(\boldsymbol{y_{p}})}$ denotes the entropy of $\boldsymbol{y_{p}}$, which is given by:

\begin{equation}
    H(\boldsymbol{y_{p}})=-\int_{\boldsymbol{y_{p}}} f(\boldsymbol{y_{p}}) \log f(\boldsymbol{y_{p}}) d \boldsymbol{y_{p}}
\end{equation}

${H(\boldsymbol{y_{p} \mid x})}$ denotes the conditional entropy of $\boldsymbol{y_{p}}$ given the known value of $\boldsymbol{x}$, and this is given by:

\begin{equation}
    H(\boldsymbol{y_{p} \mid x})=-\int_{\boldsymbol{y_{p}, x}} f(\boldsymbol{y_{p},x}) \log f(\boldsymbol{y_{p} \mid x}) d \boldsymbol{y_{p}} d \boldsymbol{x}
\end{equation}

${H(\boldsymbol{y_{p} \mid x})}$ \& ${H(\boldsymbol{y_{p}})}$, where $\boldsymbol{{y_{p}}}$ \& $\boldsymbol{x}$ are continuous variables, are entropy values that are derived through nearest neighbour entropy approximation methods from~\cite{kozachenko1987sample}. Entropy and therefore uncertainty coefficient will be determined with samples of $\boldsymbol{y_{p}}$ \& $\boldsymbol{x}$ that we obtain from their respective time series, where $\boldsymbol{x}$ is lagged one day before $\boldsymbol{y_{p}}$. 

\section{Implementation and Results}~\label{section: results}
\subsection{Correlation results for sustainable energy stocks}

\begin{figure}[b!]
   ~\centering
   ~\includegraphics[width=\linewidth,height=6cm]{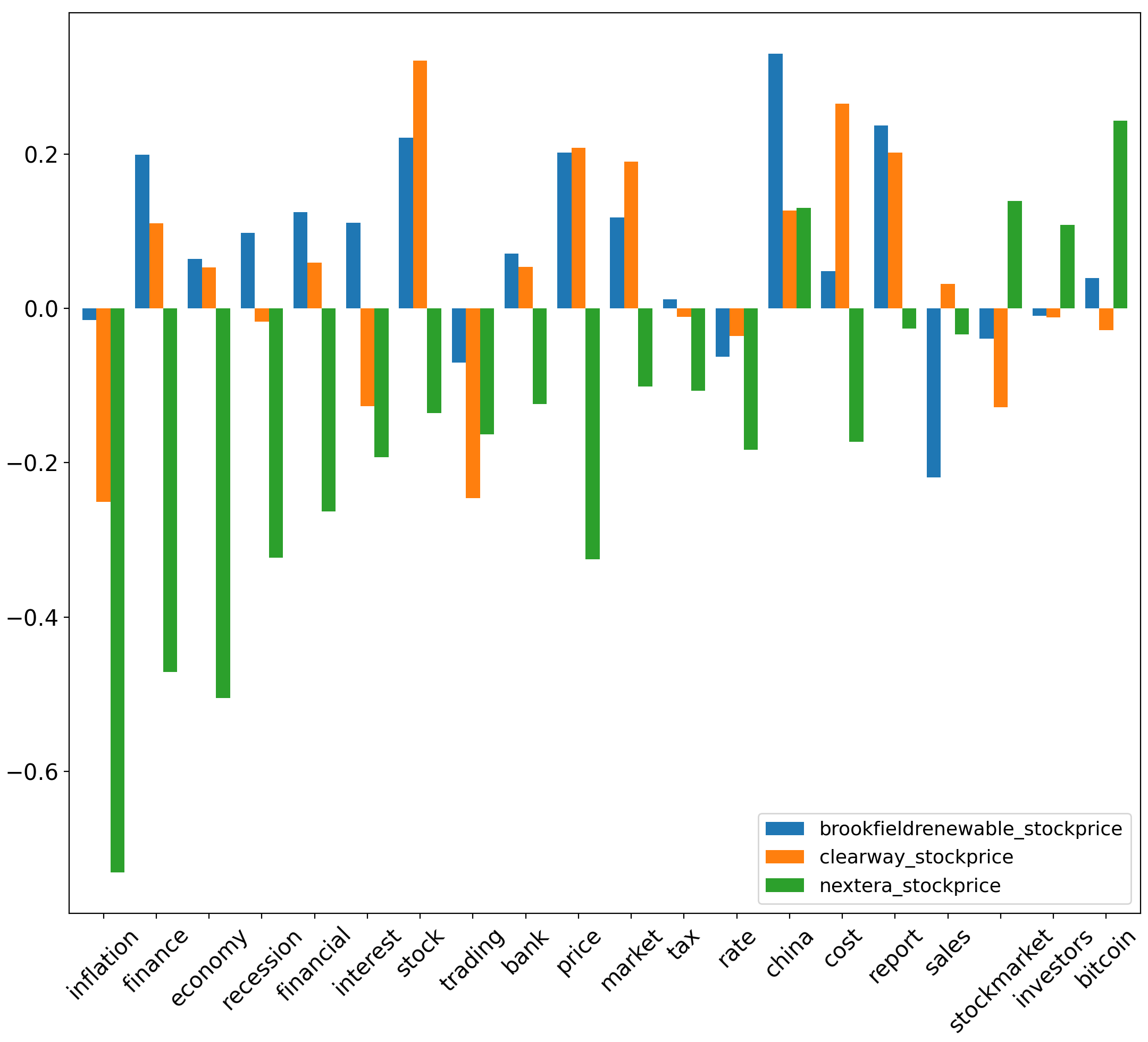}
\caption{Pearson correlation for lagged positive absolute aspect sentiment scores \&  sustainable energy stock prices}
\end{figure}

\begin{figure}[t!]
   ~\centering
   ~\includegraphics[width=\linewidth, height=6cm]{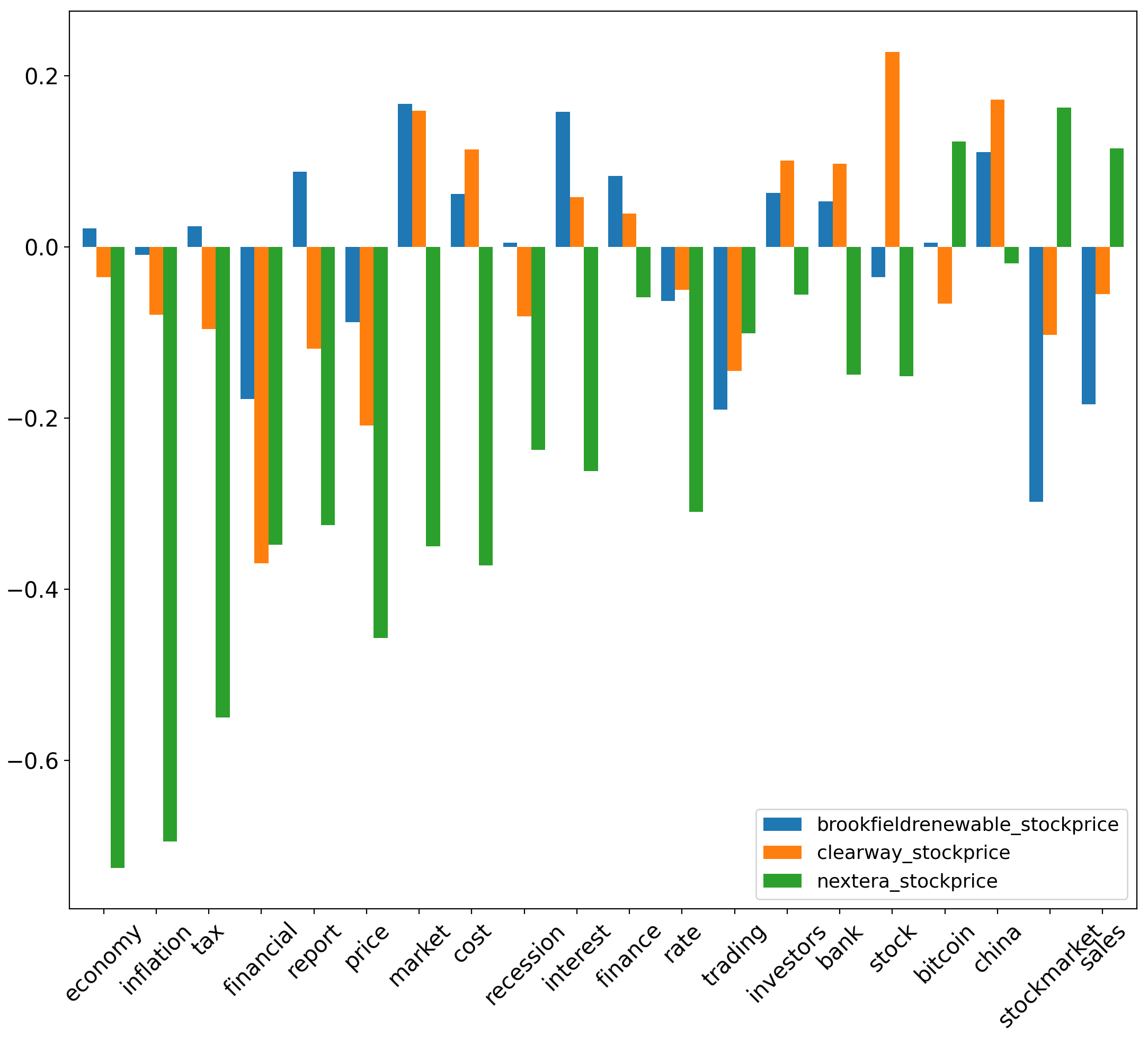}
\caption{Pearson correlation for lagged negative absolute aspect sentiment scores \&  sustainable energy stock prices}
\end{figure}

Though~\cite{li2022pearson} has considered separate thresholds for r (i.e. $0.4 <|r| < 0.6$), this paper will consider correlations of $|r| > 0.4$ to be statistically significant.
For absolute aspect sentiment scores, $y_{p,NEE}$ yields the greatest magnitude for correlation compared to other sustainable energy stocks. Specifically, $y_{p,NEE}$ yields $r$ values of $-0.731, -0.505, -0.471$ with ${x_{fp}}$ corresponding to the \textit{(inflation, economy, finance)} aspects respectively. Conversely, $y_{p,NEE}$ yields  $r$ values of $-0.726,-0.695,-0.55$ \& $-0.457$ with ${x_{fn}}$ corresponding to the respective aspects of \textit{(economy, inflation, tax, price)}. 
${r}$ values for normalised aspect sentiment scores, ${x_{nfp}}$ \& ${x_{nfn}}$, are noticeably smaller in magnitude compared to absolute aspect sentiment scores. $y_{p,NEE}$ yields an ${r}$ value of $0.44$ with ${x_{nfp}}$ for the \textit{stockmarket} aspect, and an $r$ value of $-0.449$ with ${x_{nfn}}$ for the \textit{economy} aspect. Additionally, $y_{p,CWEN}$ yields a ${r}$ value of $-0.405$ with ${x_{nfn}}$ for the \textit{financial} aspect.

\begin{figure}[b!]
   ~\centering
   ~\includegraphics[width=\linewidth,height=6cm]{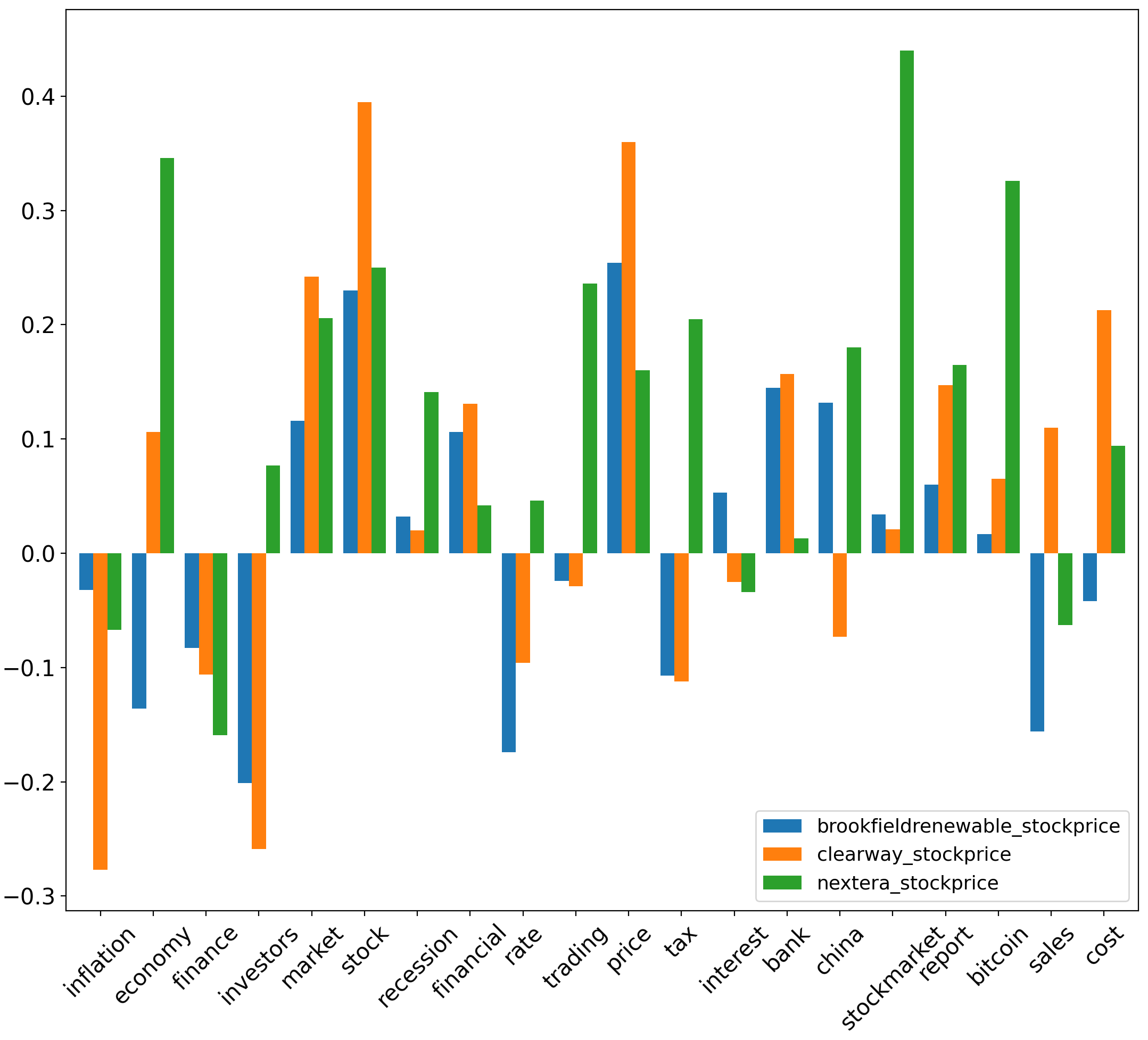}
   ~\caption{Pearson correlation for lagged positive normalised aspect sentiment scores \& sustainable energy stock prices}
\end{figure}

\begin{figure}[t!]
   ~\centering
   ~\includegraphics[width=\linewidth,height=6cm]{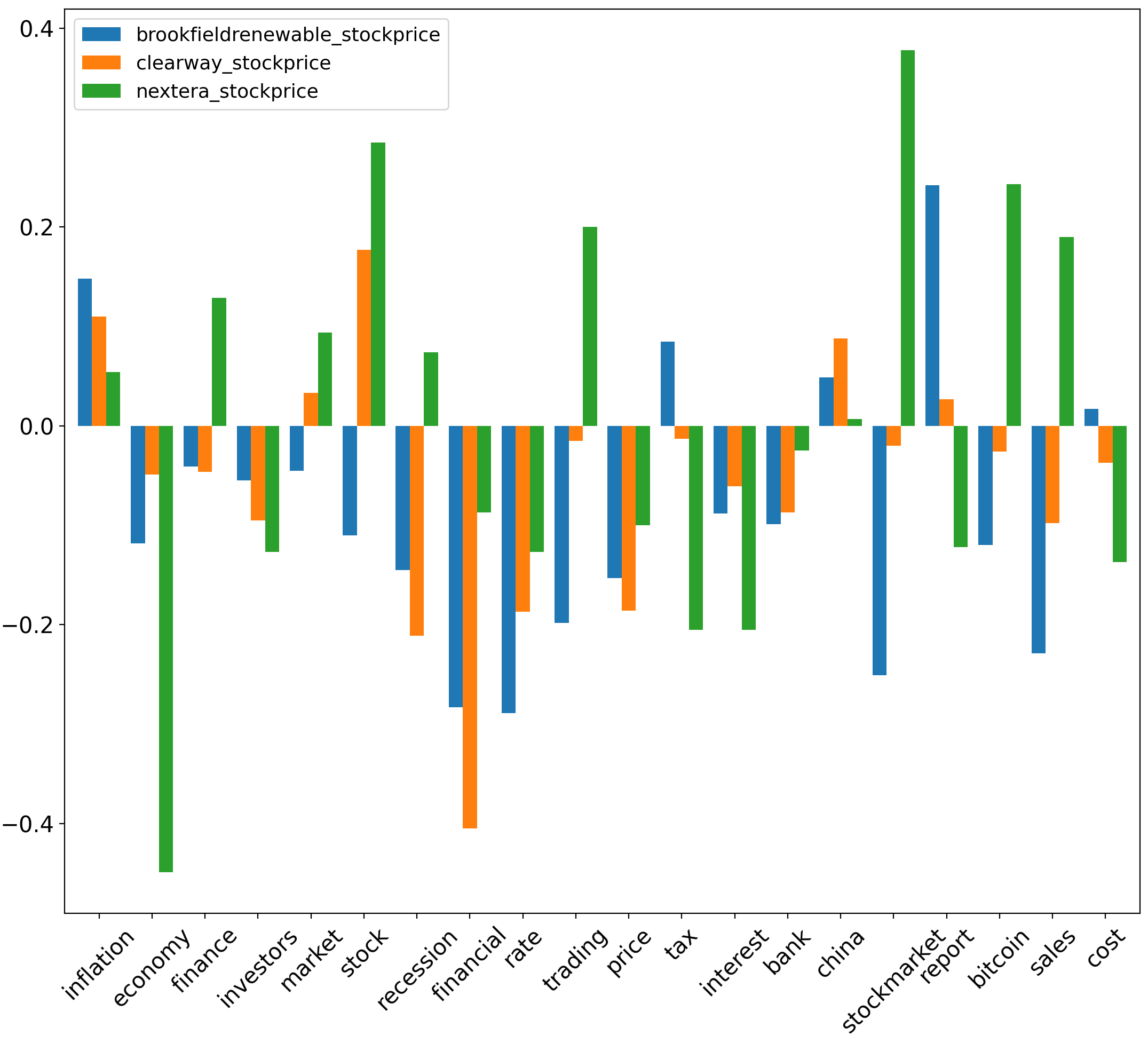}
   ~\caption{Pearson correlation for lagged negative normalised aspect sentiment scores \& sustainable energy stock prices}
\end{figure}

\begin{table}[h!]
    \centering
    \caption{Lagged aspect sentiment scores that Granger cause respective sustainable stock prices}
    \begin{tabular}{|c|c|c|}
    \hline
    Brookfieldrenewable & Clearway & Nextera \\
    \hline 
        \textit{tax ${x_{fp},x_{nfp}}$} & \textit{finance ${x_{fn}, x_{nfn}}$} & \textit{finance ${x_{fn}, x_{nfn}}$}  \\
        \textit{stock ${x_{nfn}}$} & \textit{price ${x_{fn}}$}  & \textit{investors ${x_{fn}}$}  \\
         & \textit{bitcoin ${x_{fn}}$}  & \textit{cost ${x_{fp}}$} \\
        & \textit{tax ${x_{nfp}}$} & \textit{investors ${x_{nfp}}$} \\
        & \textit{stock ${x_{nfn}}$} &  \\
    \hline    
    \end{tabular}
    \label{tab: Granger causality_aspects_sus}
\end{table}

\subsection{Granger causality results for sustainable energy stocks}

We conduct the Granger causality test to determine whether the different aspect sentiment scores, $\boldsymbol{x}$, Granger cause the various stock prices $\boldsymbol{y_{p}}$ at the 0.05 significance level. From table \ref{tab: Granger causality_aspects_sus}, we observe that more aspect sentiment scores Granger cause $y_{p,CWEN}$ \& $y_{p,NEE}$ compared to $y_{p,BEPC}$. Additionally, we observe that aspect sentiment scores pertaining to aspects \textit{(finance, tax)} Granger cause sustainable energy stock prices for the most occurrences compared to sentiment scores for other aspects. 

\subsection{Uncertainty coefficient results for sustainable energy stocks}
\begin{figure}[b!]
   ~\centering
   ~\includegraphics[width=\linewidth, height = 6cm]{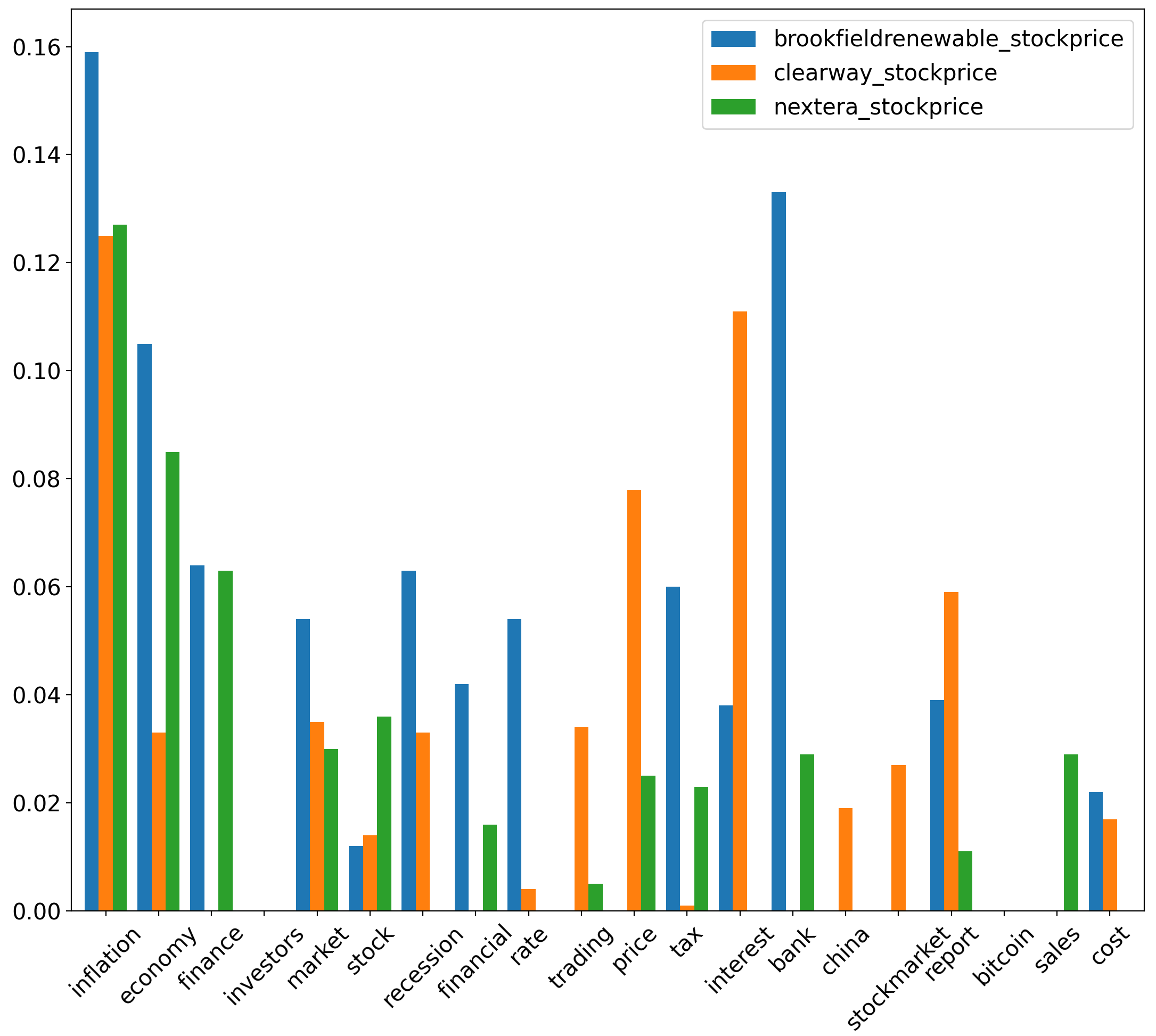}
   ~\caption{Uncertainty coefficient for lagged positive absolute aspect sentiment scores \&  sustainable energy stock prices}
   \label{fig: uncertainty_coeff_abs_sus}
\end{figure}

We highlight the highest uncertainty coefficient values between respective aspect sentiment scores and traditional energy stock prices. Specifically, from figure \ref{fig: uncertainty_coeff_abs_sus}, we observe that between ${x_{fp}}$ corresponding to the \textit{inflation} aspect \& ($y_{p,BEPC}$, $y_{p,NEE}$, $y_{p,CWEN}$), uncertainty coefficient values are relatively high at 0.159, 0.127 \& 0.125 respectively. Relative to other aspects, uncertainty coefficient also has significant value, at 0.133 between ${x_{fp}}$ corresponding to the aspect of \textit{bank} \& $y_{p,BEPC}$. Between ${x_{fp}}$ corresponding to \textit{interest} \& $y_{p,CWEN}$, it also yields a uncertainty coefficient score of 0.111.
Among all uncertainty coefficient corresponding to ${x_{fn}}$ \& sustainable energy stocks ${\boldsymbol{y_{p}}}$, uncertainty coefficient is highest between ${x_{fn}}$ corresponding to \textit{(inflation, cost, report, tax)} \& $y_{p,BEPC}$, yielding values of 0.166, 0.143, 0.125 \& 0.12 respectively. The next most significant is ${x_{fn}}$ corresponding to \textit{(inflation, economy)} \& $y_{p,NEE}$, where uncertainty coefficient values are at 0.119 \& 0.116 respectively. 

\begin{figure}[t!]
   ~\centering
   ~\includegraphics[width=\linewidth, height = 6cm]{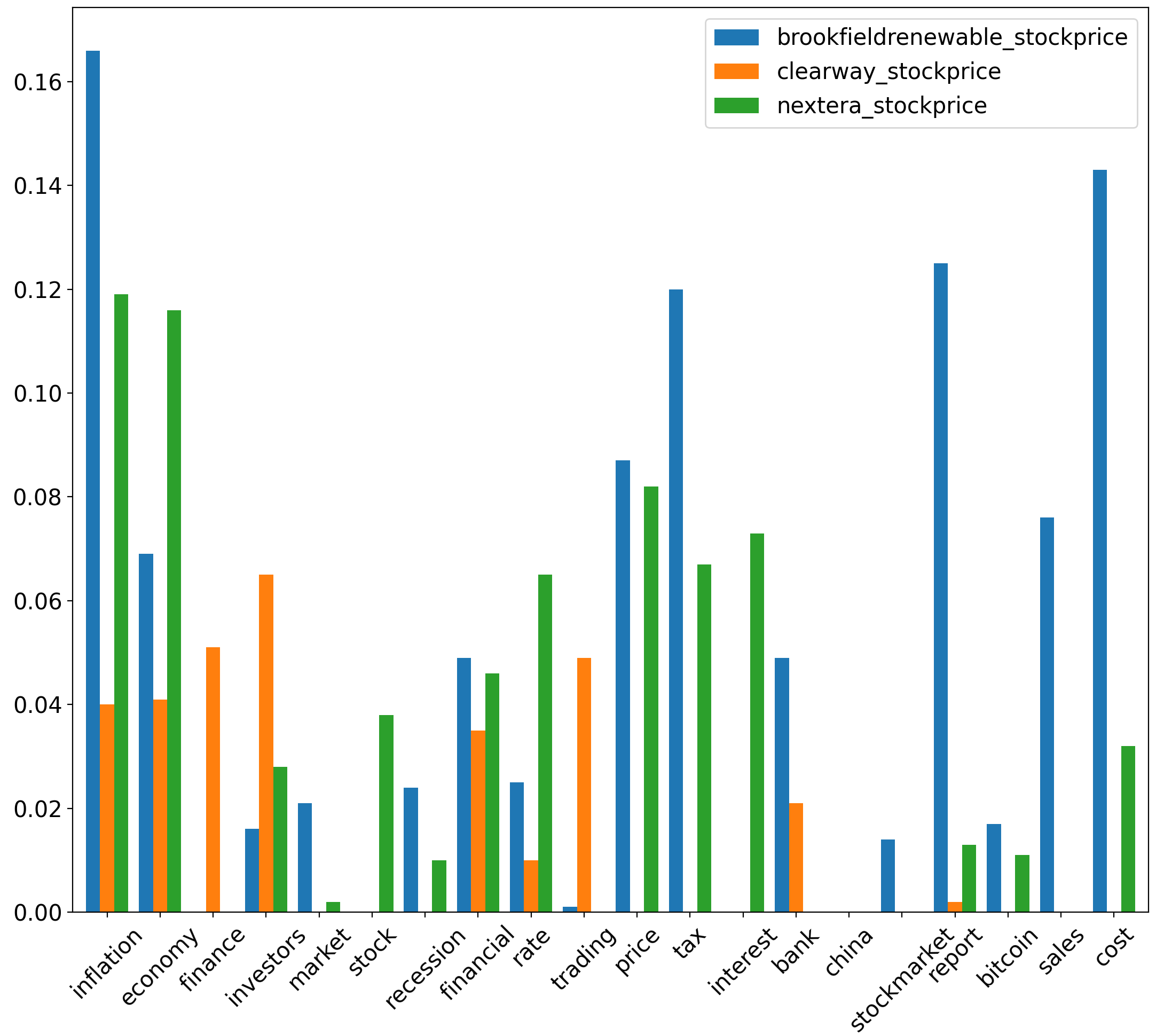}
   ~\caption{Uncertainty coefficient for lagged negative absolute aspect sentiment scores \&  sustainable energy stock prices}
   \label{fig: uncertainty_coeff_abs_sus}
\end{figure}

\begin{figure}[b!]
   ~\centering
   ~\includegraphics[width=\linewidth, height=6cm]{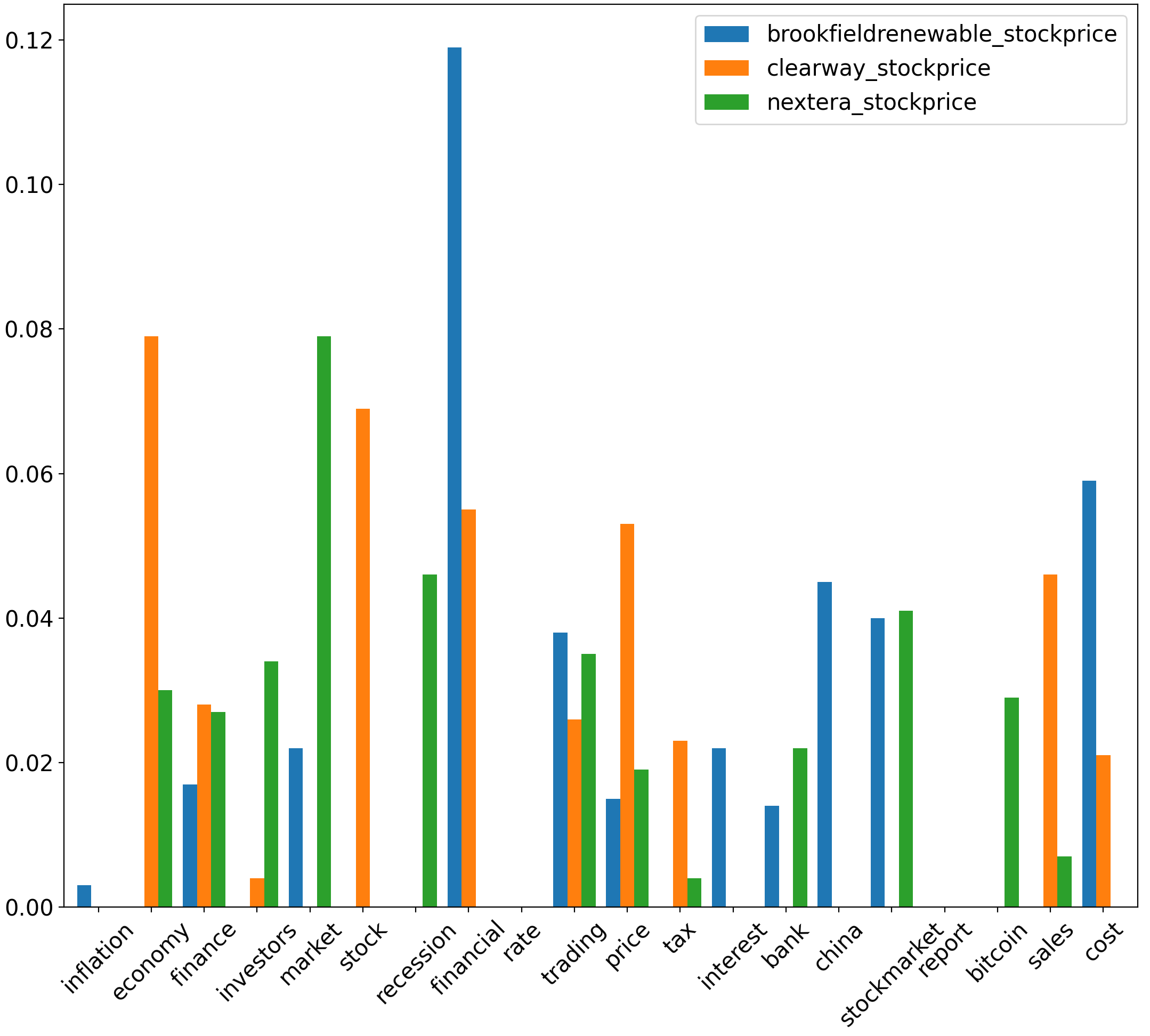}
   ~\caption{Uncertainty coefficient for lagged positive normalised aspect sentiment scores \&  sustainable energy stock prices}
\end{figure}

\begin{figure}[t!]
   ~\centering
   ~\includegraphics[width=\linewidth, height=6cm]{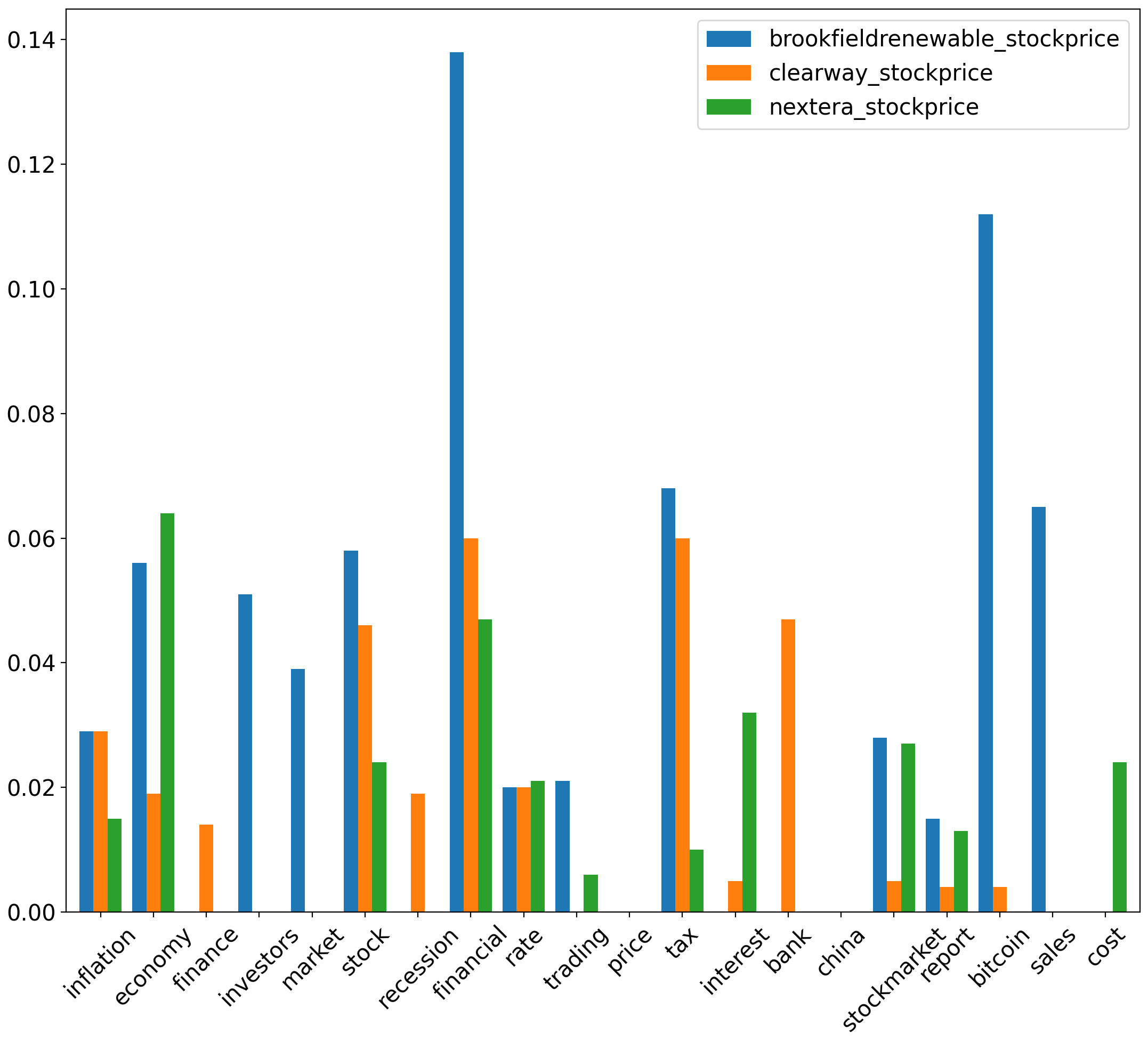}
   ~\caption{Uncertainty coefficient for lagged negative normalised aspect sentiment scores \&  sustainable energy stock prices}
   \label{fig: uncertainty_coeff_norm_sus}
\end{figure}

For uncertainty coefficient corresponding to ${x_{nfp}}$ \& sustainable energy stocks ${\boldsymbol{y_{p}}}$, the uncertainty coefficient value is highest between ${x_{nfp}}$ corresponding to the aspect \textit{financial} and $y_{p,BEPC}$, yielding a value of 0.119. Besides this, uncertainty coefficient values are also relatively significant between \textit{economy} ${x_{nfp}}$ \& $y_{p,CWEN}$, yielding a value of 0.079. Finally, uncertainty coefficient yields a value of 0.079 between \textit{market} ${x_{nfp}}$ \& $y_{p,NEE}$. 
Lastly, for uncertainty coefficients derived from ${x_{nfn}}$ \& sustainable energy stocks ${\boldsymbol{y_{p}}}$, the highest values are yielded between ${x_{nfn}}$ corresponding to $\textit{(financial, bitcoin)}$ \& $y_{p,BEPC}$ at 0.138 \& 0.112 respectively. 

\subsection{Correlation results for traditional energy stocks}
Of the traditional energy stocks analysed, British Petroleum and Shell consistently yield the greatest magnitude for Pearson correlation values.

\begin{figure}[b!]
   ~\centering
   ~\includegraphics[width=\linewidth,height=6cm]{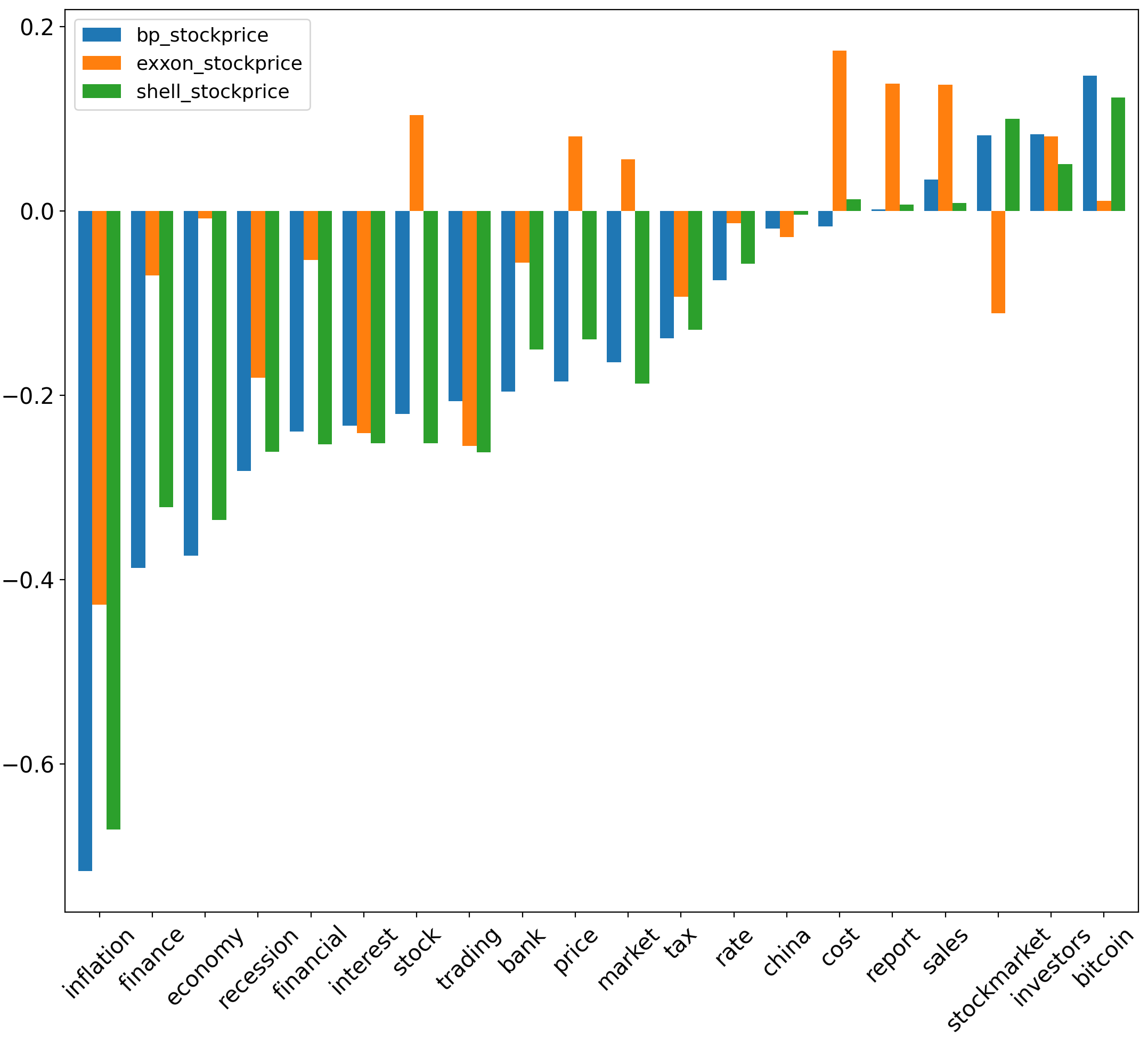}
    ~\caption{Pearson correlation for lagged positive absolute aspect sentiment scores \&  traditional energy stock prices}
\end{figure}

\begin{figure}[t!]
   ~\centering
   ~\includegraphics[width=\linewidth,height=6cm]{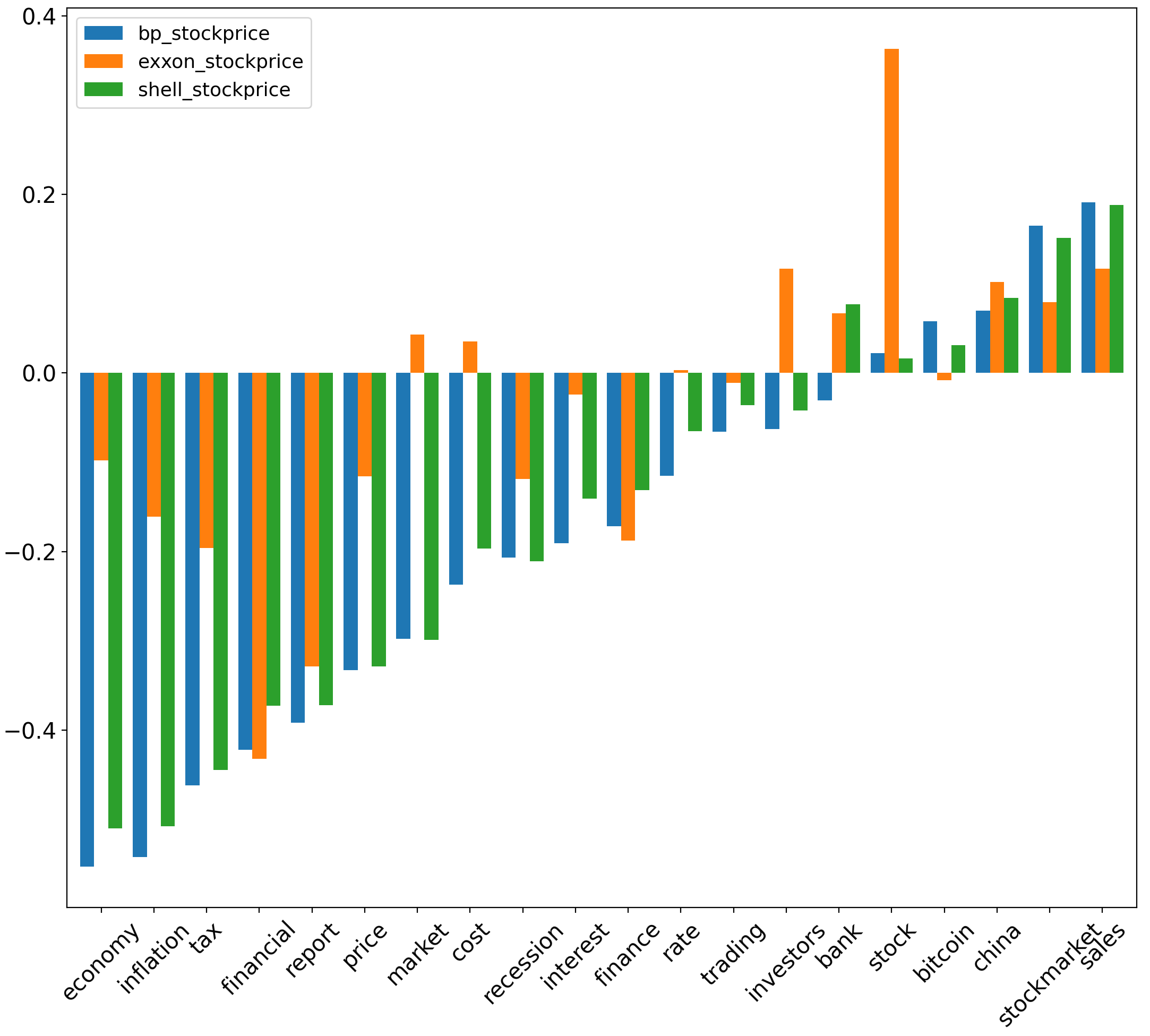}
   ~\caption{Pearson correlation for lagged negative absolute aspect sentiment scores \&  traditional energy stock prices}
   \label{fig: r_absolute_aspect_sentiment_scores}
\end{figure}

\begin{figure}[b!]
   ~\centering
   ~\includegraphics[width=\linewidth,height=6cm]{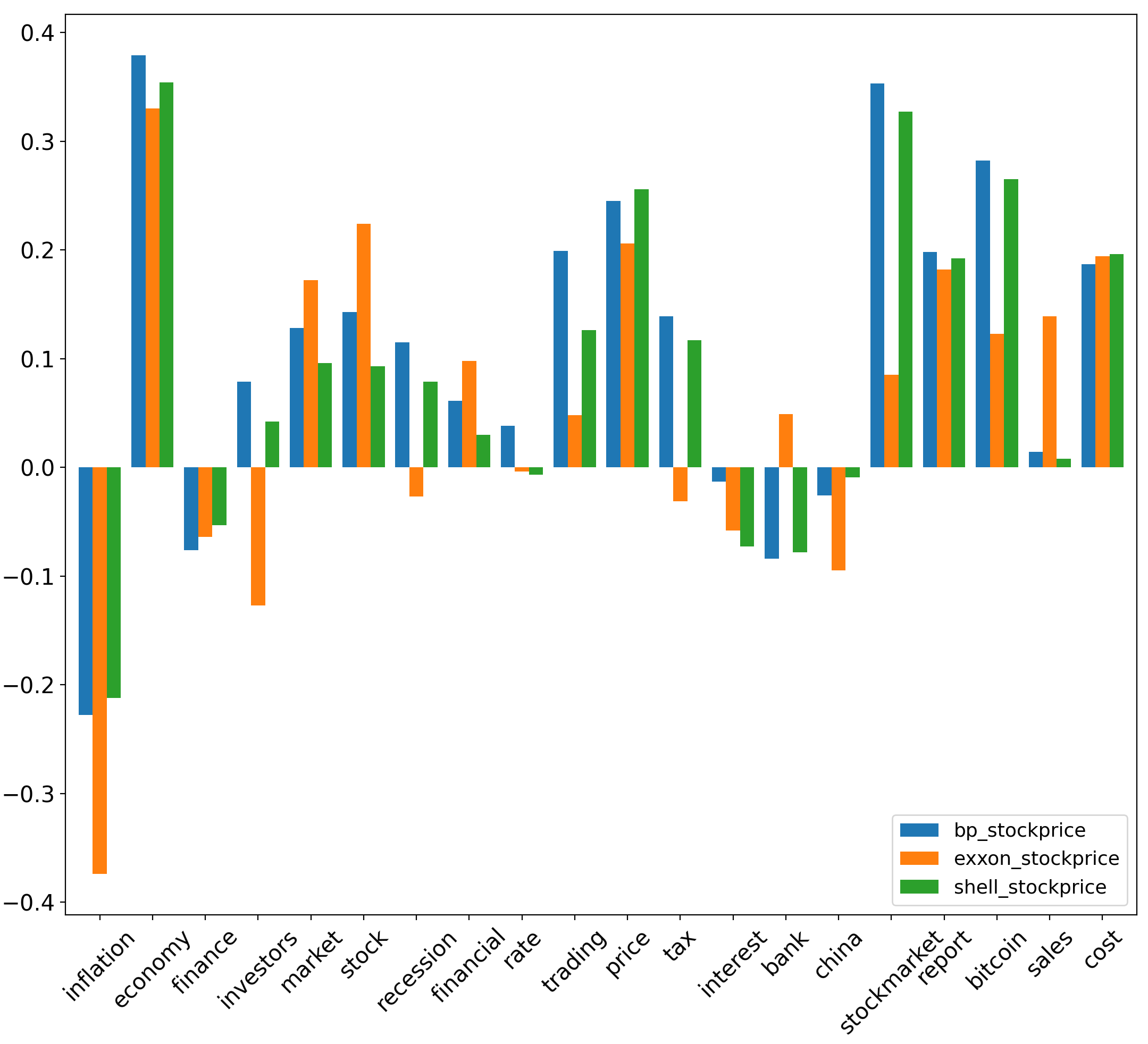}
   ~\caption{Pearson correlation values for lagged positive normalised aspect sentiment scores \&  traditional energy stock prices}\label{fig: r_normalized_aspect_sentiment_scores}
\end{figure}

\begin{figure}[t!]
   ~\centering
   ~\includegraphics[width=\linewidth,height=6cm]{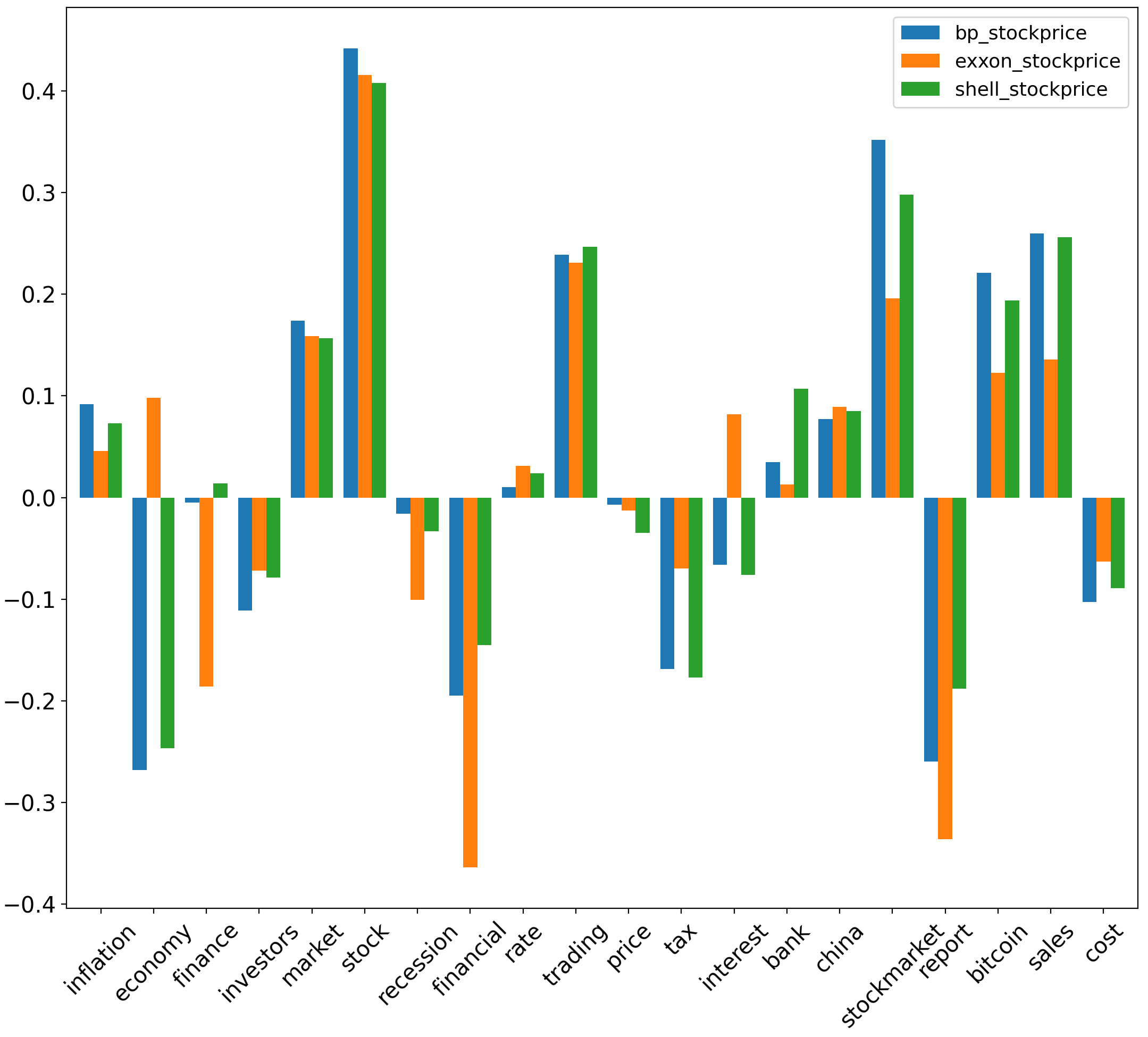}
   ~\caption{Pearson correlation for lagged negative normalised aspect sentiment scores \&  traditional energy stock prices}\label{fig: r_normalized_aspect_sentiment_scores}
\end{figure}

For absolute aspect sentiment scores, ($y_{p,BP}$, $y_{p,SHEL}$ $y_{p,XOM}$) have $r$ values of $-0.716$, $-0.671$ \& $-0.427$ with ${x_{fp}}$ corresponding to the \textit{inflation} aspect. $y_{p,BP}$ also has r values of $-0.553$, $-0.542$, $-0.462$ \& $-0.422$ with ${x_{fn}}$ corresponding to the (\textit{economy, inflation, tax, financial}) aspects respectively. Additionally, $y_{p,XOM}$ has an r value of $-0.432$ with ${x_{fn}}$ corresponding to the \textit{financial} aspect. $y_{p,SHEL}$ has r values of $-0.51$, $-0.508$ \& $-0.445$ with ${x_{fn}}$ corresponding to the (\textit{economy, inflation, tax}) aspects, respectively.
For normalised aspect sentiment scores, ($y_{p,BP}$, $y_{p,XOM}$ \& $y_{p,SHEL}$) have $r$ values of $0.442, 0.416, 0.408$ respectively with ${x_{nfn}}$ corresponding to the \textit{stock} aspect.

\begin{table}[h!]
    \caption{Lagged aspect sentiment scores that Granger Cause respective traditional stock prices}
    \centering
    \begin{tabular}{|c|c|c|}
    \hline
    BP & Exxon & Shell \\
    \hline 
        \textit{financial ${x_{fp}, x_{fn}}$} & \textit{financial ${x_{fn}, x_{nfn}}$} & \textit{stock ${x_{fp}}$}  \\
        \textit{financial ${x_{nfp}, x_{nfn}}$}
         & \textit{trading ${x_{fp}}$} &  \textit{rate ${x_{nfn}}$} \\
        \textit{recession ${x_{fp}}$} & \textit{economy ${x_{nfp}}$}  &  \\
         \textit{bitcoin ${x_{fn}}$} & \textit{stock ${x_{nfp}}$} &  \\
         \textit{stock ${x_{fp},x_{nfp},x_{nfn}}$}& \textit{cost ${x_{nfp}}$} & \\
        \textit{inflation ${x_{nfn}}$}&  &  \\
    \hline    
    \end{tabular}
    \label{tab: Granger causality_aspects_trad}
\end{table}

\subsection{Granger causality results for traditional energy stocks}

A greater number of aspect sentiment scores Granger cause traditional energy stock prices relative to sustainable energy stocks. Additionally, as highlighted in table \ref{tab: Granger causality_aspects_trad}, more aspect sentiment scores Granger cause $y_{p,BP}$ \& $y_{p,XOM}$ compared to $y_{p,SHEL}$. Moreover, aspect sentiment scores pertaining to the \textit{financial} aspect Granger cause traditional energy stock prices most frequently compared to sentiment scores corresponding to other aspects. 

\subsection{Uncertainty coefficient results for traditional energy stocks}
\begin{figure}[b!]
   ~\centering
   ~\includegraphics[width=\linewidth, height = 6cm]{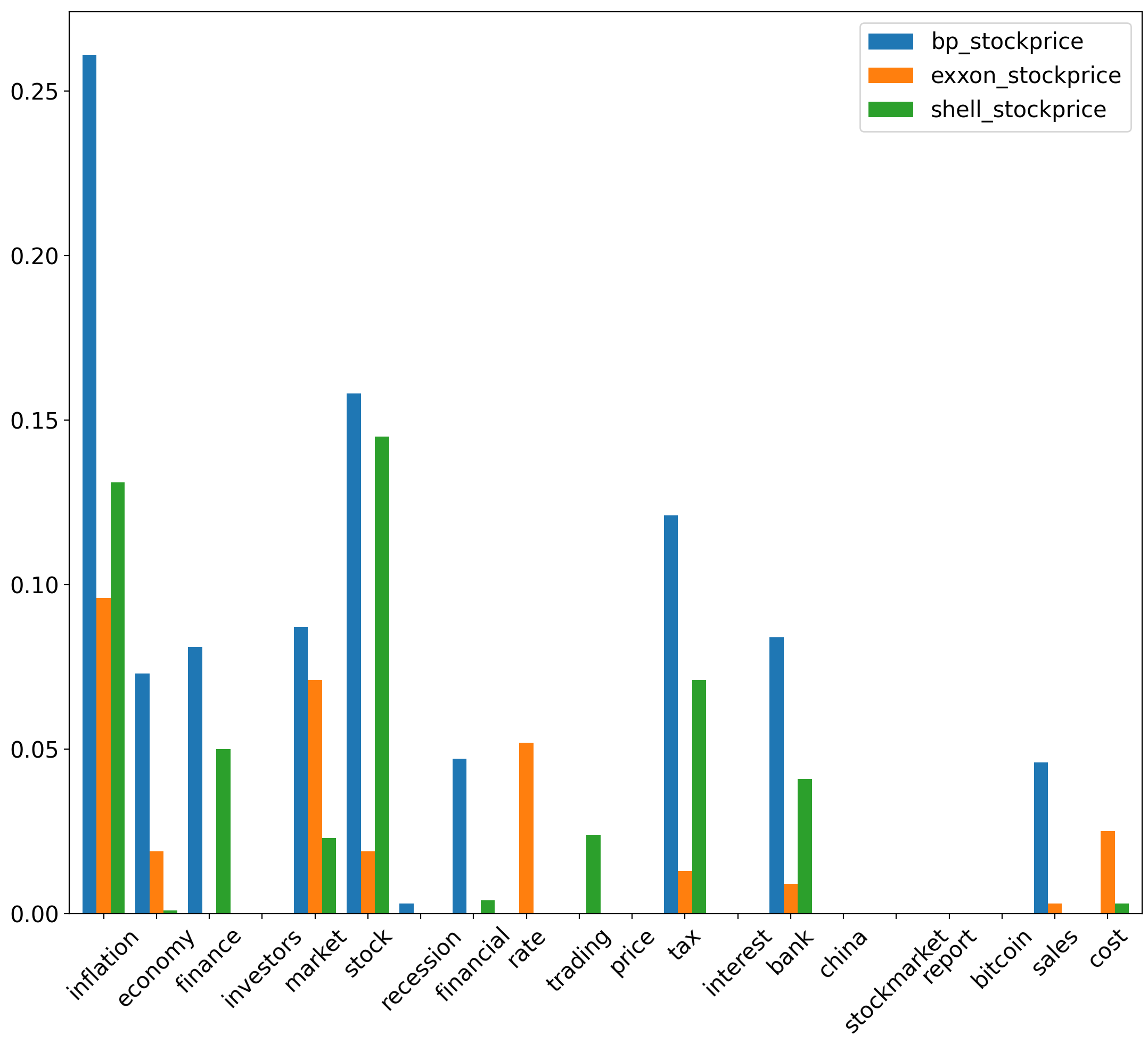}
   ~\caption{Uncertainty coefficient for lagged positive absolute aspect sentiment scores \&  traditional energy stock prices}
\end{figure}

We highlight the greatest uncertainty coefficient values between sentiment scores and stock prices for traditional energy stocks. Specifically, among all uncertainty coefficient values corresponding to ${x_{fp}}$ \& ${\boldsymbol{y_{p}}}$, uncertainty coefficient values are greatest between ${x_{fp}}$ corresponding to \textit{inflation} \& $y_{p,BP}$, at 0.261. Uncertainty coefficient values are also relatively high between ${x_{fp}}$ corresponding to \textit{stock} and ($y_{p,BP}$ \& $y_{p,XOM}$), at 0.158 \& 0.145, respectively. 
Among all uncertainty coefficient values corresponding to ${x_{fn}}$ \& ${\boldsymbol{y_{p}}}$, the uncertainty coefficient values between ${x_{fn}}$ corresponding to (\textit{economy, inflation, china, financial}) \& $y_{p,BP}$ are highest at 0.29, 0.196, 0.162 \& 0.14 respectively. Additionally, the uncertainty coefficient value between ${x_{fn}}$ corresponding to \textit{inflation} \& $y_{p,SHEL}$ is relatively high at 0.14. These uncertainty coefficient values are significant compared to those yielded between ${x_{fn}}$ corresponding to different aspects \& ${\boldsymbol{y_{p}}}$ of different traditional energy stocks.

\begin{figure}[t!]
   ~\centering
   ~\includegraphics[width=\linewidth, height = 6cm]{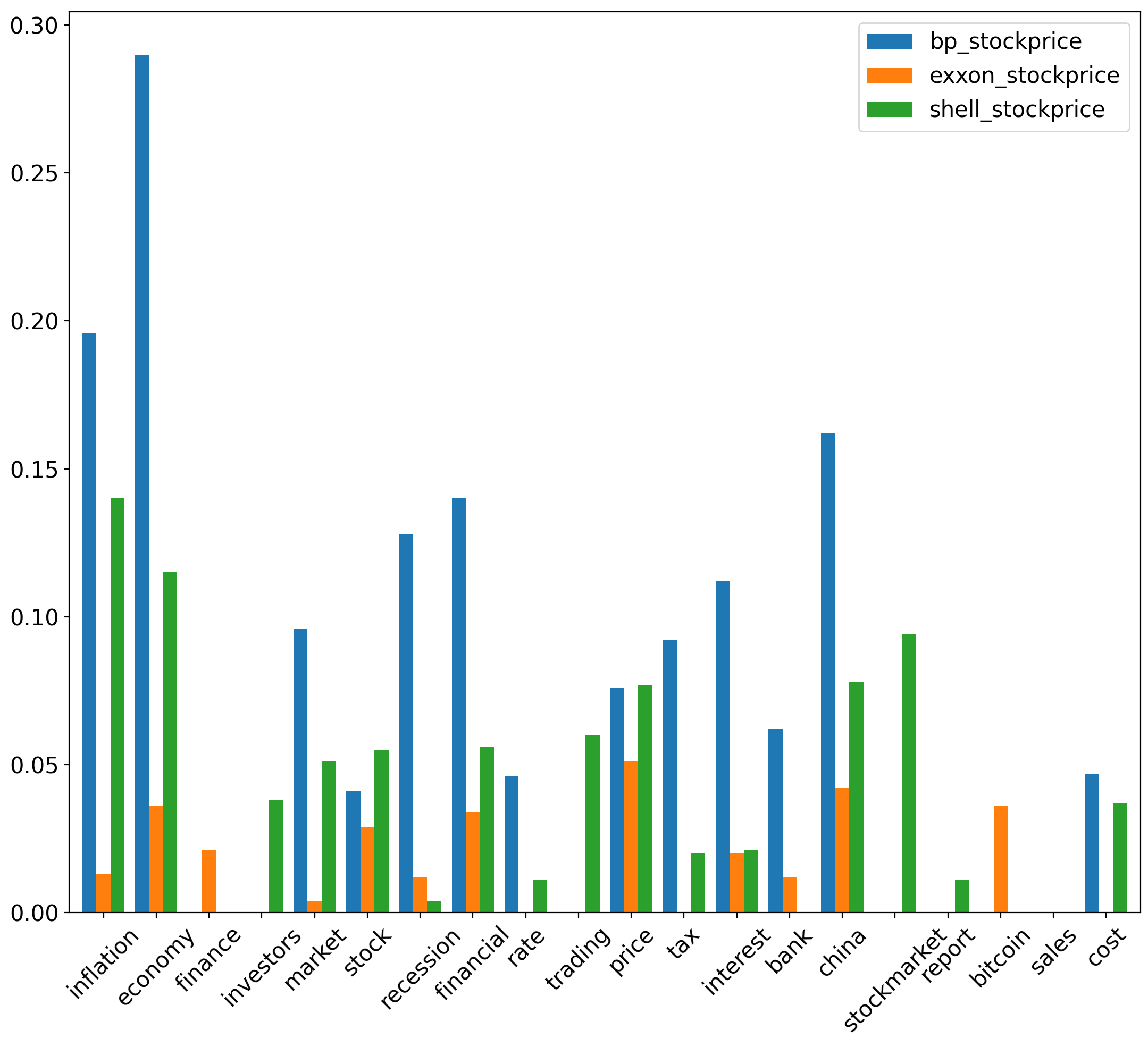}
   ~\caption{Uncertainty coefficient for lagged negative absolute aspect sentiment scores \&  traditional energy stock prices}
\end{figure}

\begin{figure}[b!]
   ~\centering
   ~\includegraphics[width=\linewidth, height=6cm]{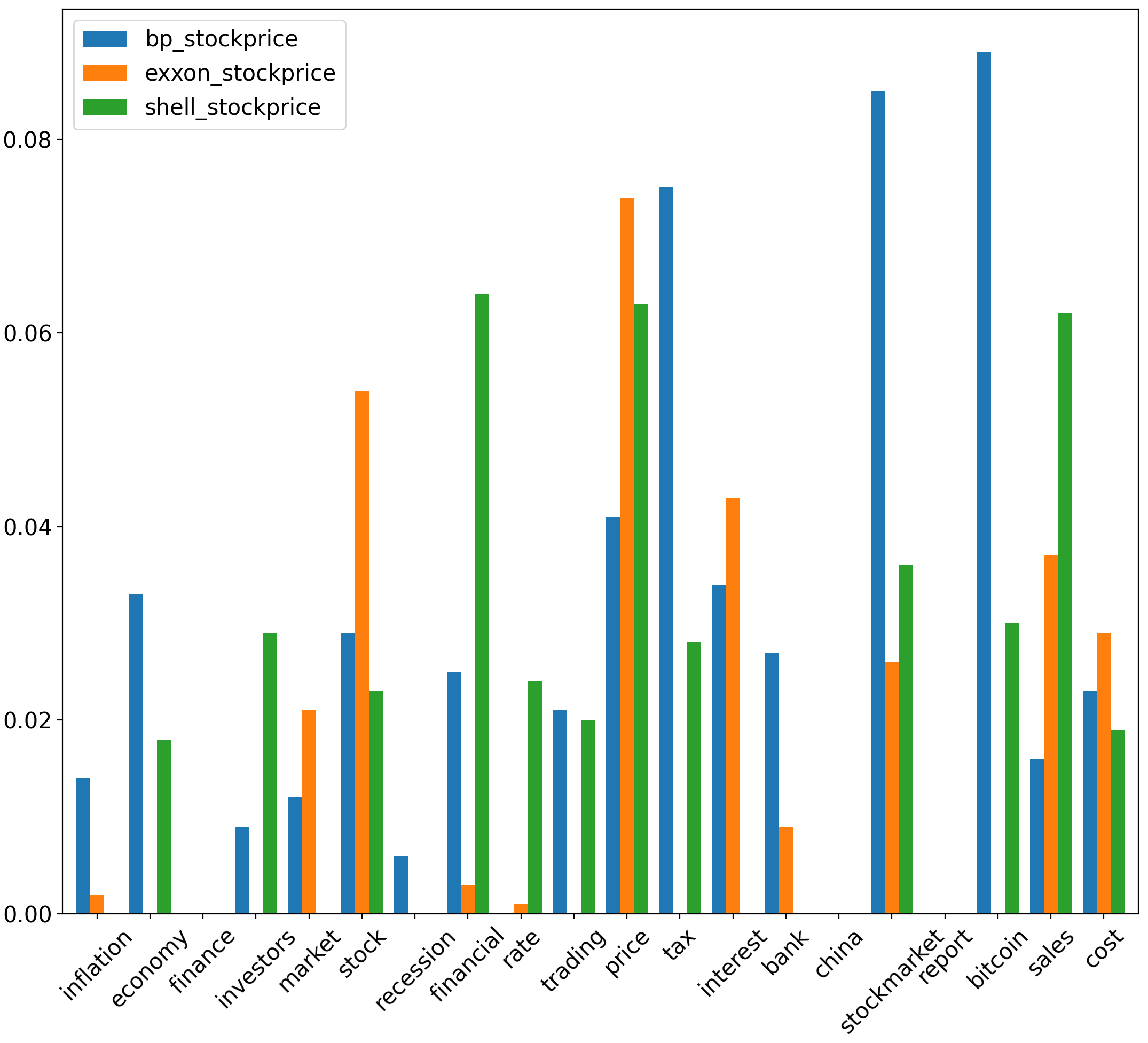}
    ~\caption{Uncertainty coefficient for lagged positive normalised aspect sentiment scores \&  traditional energy stock prices}
\end{figure}

\begin{figure}[t!]
   ~\centering
   ~\includegraphics[width=\linewidth,height=6cm]{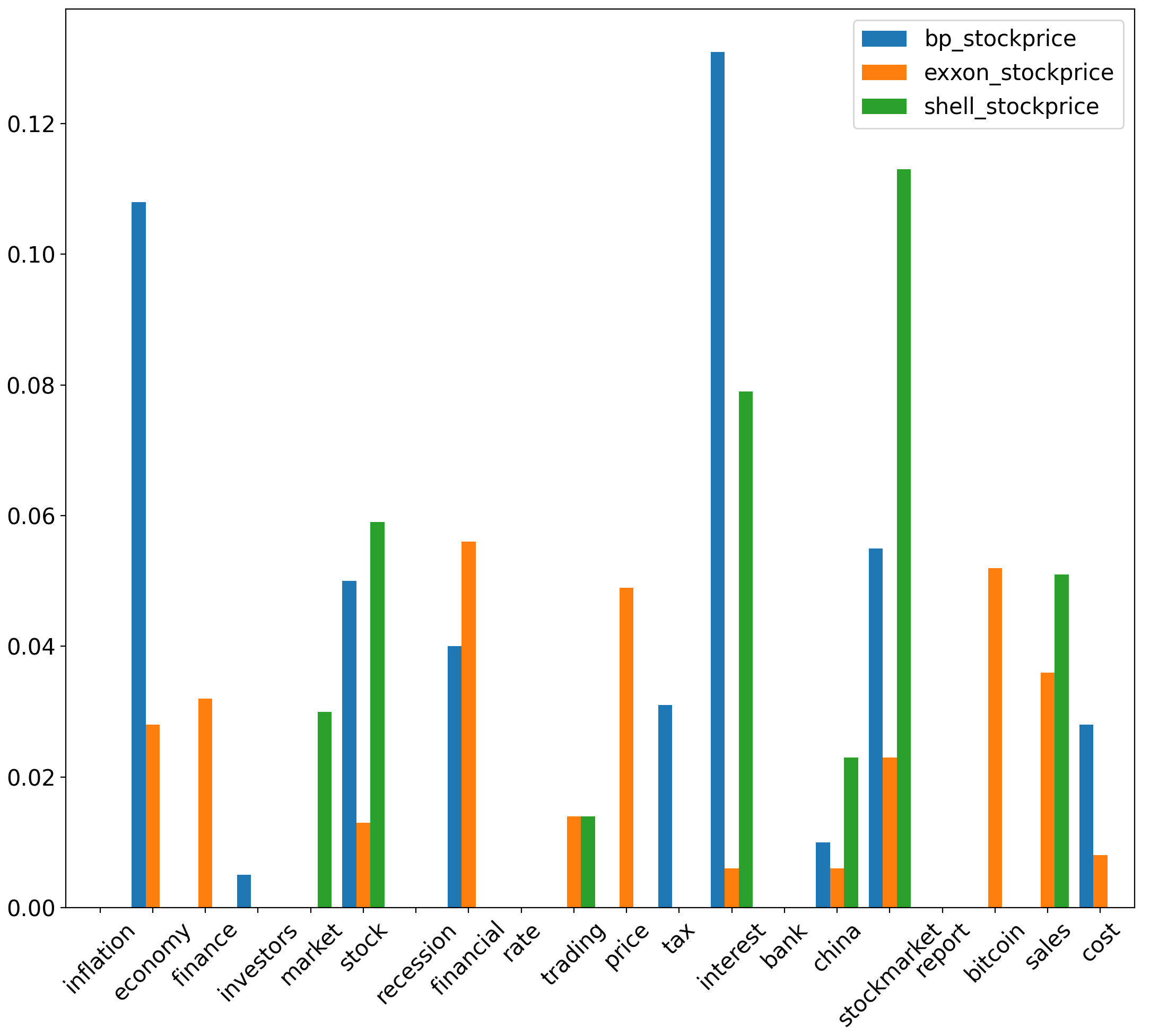}
   ~\caption{Uncertainty coefficient for lagged negative normalised aspect sentiment scores \&  traditional energy stock prices}
\end{figure}

Between ${x_{nfp}}$ corresponding to (\textit{bitcoin, stockmarket, tax}) \& $y_{p,BP}$, uncertainty coefficient values are at 0.089, 0.085 \& 0.075 respectively. Furthermore, ${x_{nfp}}$ for \textit{price} \& $y_{p,XOM}$ yields an uncertainty coefficient value of 0.074. These values are significant relative to uncertainty coefficents derived between ${x_{nfp}}$ of other aspects and different traditional energy ${\boldsymbol{y_{p}}}$.
Between ${x_{nfn}}$ of (\textit{interest, economy}) \& $y_{p,BP}$, uncertainty coefficient values are at 0.131 \& 0.108 respectively. Moreover, the uncertainty coefficient between \textit{stockmarket} ${x_{nfn}}$ \& $y_{p,SHEL}$ is at 0.113. These values are significant relative to uncertainty coefficients derived between ${x_{nfn}}$ of other aspects and various traditional energy ${\boldsymbol{y_{p}}}$.

\section{Discussion}~\label{section: discussion}
\subsection{Financially explainable correlations}\label{sec:Pearson correlation_explainable_discussion}
We will interpret financially meaningful aspects that have a clear correlation, $|r| > 0.4$, between different aspect sentiment scores, $\boldsymbol{x}$, and stock prices $\boldsymbol{y_{p}}$.
For the aspect \textit{tax}, ${x_{fn}}$ yields ${-r}$ with $y_{p,NEE}$, $y_{p,SHEL}$ \& $y_{p,BP}$. For \textit{price}, ${x_{fn}}$ yields ${-r}$ with $y_{p,NEE}$. For \textit{financial}, ${x_{fn}}$ yields ${-r}$ with ${y_{p,XOM}}$ \& ${y_{p,BP}}$, while ${x_{nfn}}$ yields
${-r}$ with $y_{p,CWEN}$. For~\textit{economy}, ${x_{fn}}$ yields $-r$ with the $y_{p,NEE}$, $y_{p,SHEL}$ \& $y_{p,BP}$, while ${x_{nfn}}$ yields significant $-r$ with $y_{p,NEE}$. In other words, a reduction in negative sentiment scores with respect to the aspects \textit{tax, price, financial, economy} is correlated with an increase in prices for separate stocks. The economic interpretation of this is that higher stock prices are correlated with a reduction in pessimism (which essentially translates to more optimism) pertaining to the different components for economic and business conditions (taxes, prices, financial situation) as well as the economy in general. Additionally, for \textit{stockmarket}, ${x_{nfp}}$ yields ${+r}$ with $y_{p,NEE}$. This indicates that greater positive sentiment about the stockmarket is correlated with a rise in nextera stock price. 

Our results are consistent with the findings of~\cite{ugurlu2021monetary},~\cite{de1990noise} \&~\cite{siganos2014facebook} which state that positive sentiments usually lead to rises in stock prices. However, our findings also add a greater level of granularity to this proven relationship, as through ABSA, we distinguished between the sentiment relating to different financial aspects. As such, these results possess the capacity to be made explainable and intelligible through economic and financial theory. To elaborate, our findings corroborate with~\cite{shu2015investor}, which describes how positive sentiment for economic and business conditions (i.e. tax, prices, financial situation), as well as the economy \& stockmarket, can pertain to higher confidence amongst investors, delineating to increases in stock prices. 

However, there are also correlation results whose interpretations are less clear. For instance, both ${x_{fp}}$ and ${x_{fn}}$ corresponding to \textit{inflation} yields $-r$ with $y_{p,NEE}$, $y_{p,SHEL}$ \& $y_{p,BP}$. As they have similar r values, these stocks might be more strongly correlated with the frequency of occurrence of the \textit{inflation} aspect as opposed to sentiment scores. Other aspects such as \textit{finance \& stock} also yield unexpected results. For \textit{finance}, ${-r}$ is yielded between ${x_{fp}}$ \& $y_{p,NEE}$, while for the \textit{stock} aspect, $y_{p,SHEL}$, $y_{p,BP}$ \& $y_{p,XOM}$ yield ${+r}$ with ${x_{nfn}}$. 
Pearson correlation shows a stronger link between aspect sentiment scores and traditional energy stocks than sustainable ones. Literature suggests Twitter sentiment has limited influence on renewable energy stock prices~\cite{reboredo2018impact}. We plan to explore if this is because the sustainable energy sector is relatively nascent and its stocks may not align as closely with financial sentiment as traditional energy stocks.

\subsection{Granger causality of sentiment scores with stock prices}\label{sec:Granger causality_robustness_discussion}
We reference Granger causality results to complement the Pearson correlation results we obtained. In addition to deriving explainable and significant correlations between various financial aspect sentiment scores and stock prices, Granger causality is indicative of the forecasting power of these sentiment scores for stock prices and confirms the interdependent relationship between them. We believe a more robust and useful analyses of correlation can be done by utilising Granger causality in tandem with Pearson correlation. Chiefly, we highlight specific aspect sentiment scores that are not only strongly correlated with stock prices, but also contain sufficient information to forecast future stock prices, highlighting an interdependent relationship. ${x_{fn}}$ corresponding to the \textit{financial} aspect Granger causes $y_{p,BP}$ \&$y_{p,XOM}$, and an explainable and significant correlation is also present between them. On the other hand, ${x_{nfn}}$ for the \textit{stock} aspect Granger causes $y_{p,BP}$ and shows significant correlation as well. 

\subsection{Interpreting the most statistically significant aspects via uncertainty coefficient}

Studies have shown that the relationship between textual sentiment and stock pricing is intricate~\cite{kearney2014textual}, and stock pricing is also driven by numerous (sentiment) features~\cite{gupta2020sentiment}. To clarify this complexity, we propose integrating ABSA with the uncertainty coefficient. We enhance interpretability by computing uncertainty coefficient. This variable measures the degree of information provided by aspect sentiment scores with regard to stock prices. As such, we aim to identify the most statistically significant aspect sentiment scores related to stock prices.
Our findings indicate that uncertainty coefficient values are higher between absolute aspect sentiment scores and stock prices as compared to normalised sentiment scores, suggesting absolute sentiment scores offer more insight into stock prices. Additionally, uncertainty coefficient values are higher for traditional energy stocks compared to sustainable ones.

Next, we analyze the uncertainty coefficient values to assess which aspect sentiment scores offer the most valuable information regarding stock prices. Among all aspect sentiment scores and stock prices, the highest uncertainty coefficient is observed between ${x_{fn}}$ related to the \textit{economy} and $y_{p,BP}$. Following closely is the uncertainty coefficient between ${x_{fp}}$ for \textit{inflation} and $y_{p,BP}$. Furthermore, for $y_{p,BEPC}$, $y_{p,CWEN}$, and $y_{p,NEE}$, sentiment scores related to \textit{inflation} exhibit the highest uncertainty coefficient values. Lastly, when focusing on $y_{p,SHEL}$, sentiment scores for \textit{economy} also possess among the highest uncertainty coefficients compared to other aspects.

In summary, sentiment scores concerning the \textit{economy} and \textit{inflation} consistently contain the most informative signals regarding various stock prices. Relative to sentiment scores for other aspects, they possess the greatest uncertainty coefficient values with respect to stock prices. Additionally, sentiment scores for other aspects, such as \textit{financial, china, stockmarket, interest, stock, tax, price, bitcoin, cost, bank, market}, also exhibit elevated uncertainty coefficient values with stock prices. Notably, sentiment scores for \textit{financial} show not only high uncertainty coefficient values but also a strong and explainable correlation and Granger causality with stock prices. 

\section{Conclusion \& Future Work}~\label{section: conclusion}
This paper introduces an explainable financial analysis method using aspect-based sentiment analysis, Pearson coefficient, Granger causality, and uncertainty coefficient. It showcases the enhanced explainability and robustness achieved by integrating these statistical methods with aspect based sentiment analysis and stock prices.

The study has limitations, including its brief duration, focus on energy stocks, and use of generic social media data. Future endeavors will lengthen the study, target company-specific data, and delve into non-linear dynamics using interpretable neural networks. We also aim to incorporate microtext normalization~\cite{satapathy2020review}, text mining, emotion metrics~\cite{duong2023wildfires}, and advanced interpretability methods~\cite{mengaldo2023}. Recent strides in neurosymbolic AI for sentiment analysis~\cite{xing2023guest},~\cite{cambria2022senticnet}, also offer promising avenues for enhancing explainability in financial AI applications.

\appendices

\bibliographystyle{IEEEtran}
\bibliography{main.bib}

\begin{thebibliography}{10}
\providecommand{\url}[1]{#1}
\csname url@samestyle\endcsname
\providecommand{\newblock}{\relax}
\providecommand{\bibinfo}[2]{#2}
\providecommand{\BIBentrySTDinterwordspacing}{\spaceskip=0pt\relax}
\providecommand{\BIBentryALTinterwordstretchfactor}{4}
\providecommand{\BIBentryALTinterwordspacing}{\spaceskip=\fontdimen2\font plus
\BIBentryALTinterwordstretchfactor\fontdimen3\font minus
  \fontdimen4\font\relax}
\providecommand{\BIBforeignlanguage}[2]{{%
\expandafter\ifx\csname l@#1\endcsname\relax
\typeout{** WARNING: IEEEtran.bst: No hyphenation pattern has been}%
\typeout{** loaded for the language `#1'. Using the pattern for}%
\typeout{** the default language instead.}%
\else
\language=\csname l@#1\endcsname
\fi
#2}}
\providecommand{\BIBdecl}{\relax}
\BIBdecl

\bibitem{rudin2019stop}
C.~Rudin, ``Stop explaining black box machine learning models for high stakes
  decisions and use interpretable models instead,'' \emph{Nature machine
  intelligence}, vol.~1, no.~5, pp. 206--215, 2019.

\bibitem{liang2022aspect}
B.~Liang, H.~Su, L.~Gui, E.~Cambria, and R.~Xu, ``Aspect-based sentiment
  analysis via affective knowledge enhanced graph convolutional networks,''
  \emph{Knowledge-Based Systems}, vol. 235, p. 107643, 2022.

\bibitem{onesta}
L.~Oneto, F.~Bisio, E.~Cambria, and D.~Anguita, ``Statistical learning theory
  and {ELM} for big social data analysis,'' \emph{{IEEE} Computational
  Intelligence Magazine}, vol.~11, no.~3, pp. 45--55, 2016.

\bibitem{camcomshort}
E.~Cambria, A.~Hussain, C.~Havasi, and C.~Eckl, ``Common sense computing: From
  the society of mind to digital intuition and beyond,'' in \emph{Biometric ID
  Management and Multimodal Communication}, ser. Lecture Notes in Computer
  Science.\hskip 1em plus 0.5em minus 0.4em\relax Berlin Heidelberg: Springer,
  2009, vol. 5707, pp. 252--259.

\bibitem{xincog}
F.~Xing, F.~Pallucchini, and E.~Cambria, ``Cognitive-inspired domain adaptation
  of sentiment lexicons,'' \emph{Information Processing and Management},
  vol.~56, no.~3, pp. 554--564, 2019.

\bibitem{shortcammed}
E.~Cambria, T.~Mazzocco, A.~Hussain, and C.~Eckl, ``Sentic medoids: Organizing
  affective common sense knowledge in a multi-dimensional vector space,'' ser.
  Lecture Notes in Computer Science.\hskip 1em plus 0.5em minus 0.4em\relax
  Berlin Heidelberg: Springer-Verlag, 2011, vol. 6677, pp. 601--610.

\bibitem{camsta}
E.~Cambria, B.~Schuller, B.~Liu, H.~Wang, and C.~Havasi, ``Statistical
  approaches to concept-level sentiment analysis,'' \emph{{IEEE} Intelligent
  Systems}, vol.~28, no.~3, pp. 6--9, 2013.

\bibitem{valcon}
A.~Valdivia, V.~Luz{\'o}n, E.~Cambria, and F.~Herrera, ``Consensus vote models
  for detecting and filtering neutrality in sentiment analysis,''
  \emph{Information Fusion}, vol.~44, pp. 126--135, 2018.

\bibitem{camsev}
E.~Cambria, R.~Mao, M.~Chen, Z.~Wang, and S.-B. Ho, ``Seven pillars for the
  future of {AI},'' \emph{{IEEE} Intelligent Systems}, vol.~38, no.~6, 2023.

\bibitem{mao2021bridging}
R.~Mao and X.~Li, ``Bridging towers of multi-task learning with a gating
  mechanism for aspect-based sentiment analysis and sequential metaphor
  identification,'' in \emph{Proceedings of the AAAI conference on artificial
  intelligence}, vol.~35, no.~15, 2021, pp. 13\,534--13\,542.

\bibitem{he2022meta}
K.~He, R.~Mao, T.~Gong, C.~Li, and E.~Cambria, ``Meta-based self-training and
  re-weighting for aspect-based sentiment analysis,'' \emph{IEEE Transactions
  on Affective Computing}, vol.~15, 2024.

\bibitem{jangid2018aspect}
H.~Jangid, S.~Singhal, R.~R. Shah, and R.~Zimmermann, ``Aspect-based financial
  sentiment analysis using deep learning,'' in \emph{Companion Proceedings of
  the The Web Conference 2018}, 2018, pp. 1961--1966.

\bibitem{yang2018financial}
S.~Yang, J.~Rosenfeld, and J.~Makutonin, ``Financial aspect-based sentiment
  analysis using deep representations,'' \emph{arXiv preprint
  arXiv:1808.07931}, 2018.

\bibitem{man2019financial}
X.~Man, T.~Luo, and J.~Lin, ``Financial sentiment analysis ({FSA}): A survey,''
  in \emph{2019 IEEE International Conference on Industrial Cyber Physical
  Systems (ICPS)}.\hskip 1em plus 0.5em minus 0.4em\relax IEEE, 2019, pp.
  617--622.

\bibitem{ma2023multi}
Y.~Ma, R.~Mao, Q.~Lin, P.~Wu, and E.~Cambria, ``Multi-source aggregated
  classification for stock price movement prediction,'' \emph{Information
  Fusion}, vol.~91, pp. 515--528, 2023.

\bibitem{wang2023learning}
Z.~Wang, Z.~Hu, F.~Li, S.-B. Ho, and E.~Cambria, ``Learning-based stock
  trending prediction by incorporating technical indicators and social media
  sentiment,'' \emph{Cognitive Computation}, vol.~15, no.~3, pp. 1092--1102,
  2023.

\bibitem{picasso2019technical}
A.~Picasso, S.~Merello, Y.~Ma, L.~Oneto, and E.~Cambria, ``Technical analysis
  and sentiment embeddings for market trend prediction,'' \emph{Expert Systems
  with Applications}, vol. 135, pp. 60--70, 2019.

\bibitem{gite2021explainable}
S.~Gite, H.~Khatavkar, K.~Kotecha, S.~Srivastava, P.~Maheshwari, and N.~Pandey,
  ``Explainable stock prices prediction from financial news articles using
  sentiment analysis,'' \emph{PeerJ Computer Science}, vol.~7, p. e340, 2021.

\bibitem{luo2018beyond}
L.~Luo, X.~Ao, F.~Pan, J.~Wang, T.~Zhao, N.~Yu, and Q.~He, ``Beyond polarity:
  Interpretable financial sentiment analysis with hierarchical query-driven
  attention.'' in \emph{IJCAI}, 2018, pp. 4244--4250.

\bibitem{loginova2021forecasting}
E.~Loginova, W.~K. Tsang, G.~van Heijningen, L.-P. Kerkhove, and D.~F. Benoit,
  ``Forecasting directional bitcoin price returns using aspect-based sentiment
  analysis on online text data,'' \emph{Machine Learning}, pp. 1--24, 2021.

\bibitem{ruan2018using}
Y.~Ruan, A.~Durresi, and L.~Alfantoukh, ``Using twitter trust network for stock
  market analysis,'' \emph{Knowledge-Based Systems}, vol. 145, pp. 207--218,
  2018.

\bibitem{lee2014granger}
T.-H. Lee and W.~Yang, ``Granger-causality in quantiles between financial
  markets: Using copula approach,'' \emph{International Review of Financial
  Analysis}, vol.~33, pp. 70--78, 2014.

\bibitem{chu2016nonlinear}
X.~Chu, C.~Wu, and J.~Qiu, ``A nonlinear granger causality test between stock
  returns and investor sentiment for chinese stock market: a wavelet-based
  approach,'' \emph{Applied Economics}, vol.~48, no.~21, pp. 1915--1924, 2016.

\bibitem{hamraoui2022impact}
I.~Hamraoui and A.~Boubaker, ``Impact of twitter sentiment on stock price
  returns,'' \emph{Social Network Analysis and Mining}, vol.~12, no.~1, p.~28,
  2022.

\bibitem{kim2021information}
K.~Kim, D.~Ryu, and H.~Yang, ``Information uncertainty, investor sentiment, and
  analyst reports,'' \emph{International Review of Financial Analysis},
  vol.~77, p. 101835, 2021.

\bibitem{birru2022sentiment}
J.~Birru and T.~Young, ``Sentiment and uncertainty,'' \emph{Journal of
  Financial Economics}, vol. 146, no.~3, pp. 1148--1169, 2022.

\bibitem{satapathy2017subjectivity}
R.~Satapathy, I.~Chaturvedi, E.~Cambria, S.~S. Ho, and J.~C. Na, ``Subjectivity
  detection in nuclear energy tweets,'' \emph{Computaci{\'o}n y Sistemas},
  vol.~21, no.~4, pp. 657--664, 2017.

\bibitem{dhegra}
D.~Rajagopal, E.~Cambria, D.~Olsher, and K.~Kwok, ``A graph-based approach to
  commonsense concept extraction and semantic similarity detection,'' in
  \emph{{WWW}}, 2013, pp. 565--570.

\bibitem{salunkhe2019aspect}
A.~Salunkhe and S.~Mhaske, ``Aspect based sentiment analysis on financial data
  using transferred learning approach using pre-trained bert and regressor
  model,'' \emph{Int. Res. J. Eng. Technol.(IRJET)}, vol.~6, pp. 1097--1101,
  2019.

\bibitem{el2016learning}
M.~El-Haj, P.~E. Rayson, S.~E. Young, M.~Walker, A.~Moore, V.~Athanasakou, and
  T.~Schleicher, ``Learning tone and attribution for financial text mining,''
  2016.

\bibitem{chen2017comparative}
Y.~Chen, R.~M. Rabbani, A.~Gupta, and M.~J. Zaki, ``Comparative text analytics
  via topic modeling in banking,'' in \emph{2017 IEEE Symposium Series on
  Computational Intelligence (SSCI)}.\hskip 1em plus 0.5em minus 0.4em\relax
  IEEE, 2017, pp. 1--8.

\bibitem{smailovic2013predictive}
J.~Smailovi{\'c}, M.~Gr{\v{c}}ar, N.~Lavra{\v{c}}, and
  M.~{\v{Z}}nidar{\v{s}}i{\v{c}}, ``Predictive sentiment analysis of tweets: A
  stock market application,'' in \emph{Human-Computer Interaction and Knowledge
  Discovery in Complex, Unstructured, Big Data: Third International Workshop,
  HCI-KDD 2013, Held at SouthCHI 2013, Maribor, Slovenia, July 1-3, 2013.
  Proceedings}.\hskip 1em plus 0.5em minus 0.4em\relax Springer, 2013, pp.
  77--88.

\bibitem{granger1969investigating}
C.~W. Granger, ``Investigating causal relations by econometric models and
  cross-spectral methods,'' \emph{Econometrica: journal of the Econometric
  Society}, pp. 424--438, 1969.

\bibitem{hiemstra1994testing}
C.~Hiemstra and J.~D. Jones, ``Testing for linear and nonlinear granger
  causality in the stock price-volume relation,'' \emph{The Journal of
  Finance}, vol.~49, no.~5, pp. 1639--1664, 1994.

\bibitem{abhyankar1997uncovering}
A.~Abhyankar, L.~S. Copeland, and W.~Wong, ``Uncovering nonlinear structure in
  real-time stock-market indexes: the s\&p 500, the dax, the nikkei 225, and
  the ftse-100,'' \emph{Journal of Business \& Economic Statistics}, vol.~15,
  no.~1, pp. 1--14, 1997.

\bibitem{abramowitz1970handbook}
M.~Abramowitz and I.~Stegun, ``Handbook of mathematical functions with
  formulas, graphs, and mathematical tables'' edited by dover publications,''
  \emph{Inc., New York, Ninth Printing}, 1970.

\bibitem{kozachenko1987sample}
L.~F. Kozachenko and N.~N. Leonenko, ``Sample estimate of the entropy of a
  random vector,'' \emph{Problemy Peredachi Informatsii}, vol.~23, no.~2, pp.
  9--16, 1987.

\bibitem{li2022pearson}
G.~Li, A.~Zhang, Q.~Zhang, D.~Wu, and C.~Zhan, ``Pearson correlation
  coefficient-based performance enhancement of broad learning system for stock
  price prediction,'' \emph{IEEE Transactions on Circuits and Systems II:
  Express Briefs}, vol.~69, no.~5, pp. 2413--2417, 2022.

\bibitem{ugurlu2021monetary}
E.~Ugurlu-Yildirim, B.~Kocaarslan, and B.~M. Ordu-Akkaya, ``Monetary policy
  uncertainty, investor sentiment, and us stock market performance: New
  evidence from nonlinear cointegration analysis,'' \emph{International Journal
  of Finance \& Economics}, vol.~26, no.~2, pp. 1724--1738, 2021.

\bibitem{de1990noise}
J.~B. De~Long, A.~Shleifer, L.~H. Summers, and R.~J. Waldmann, ``Noise trader
  risk in financial markets,'' \emph{Journal of political Economy}, vol.~98,
  no.~4, pp. 703--738, 1990.

\bibitem{siganos2014facebook}
A.~Siganos, E.~Vagenas-Nanos, and P.~Verwijmeren, ``Facebook's daily sentiment
  and international stock markets,'' \emph{Journal of Economic Behavior \&
  Organization}, vol. 107, pp. 730--743, 2014.

\bibitem{shu2015investor}
H.-C. Shu and J.-H. Chang, ``Investor sentiment and financial market
  volatility,'' \emph{Journal of Behavioral Finance}, vol.~16, no.~3, pp.
  206--219, 2015.

\bibitem{reboredo2018impact}
J.~C. Reboredo and A.~Ugolini, ``The impact of twitter sentiment on renewable
  energy stocks,'' \emph{Energy economics}, vol.~76, pp. 153--169, 2018.

\bibitem{kearney2014textual}
C.~Kearney and S.~Liu, ``Textual sentiment in finance: A survey of methods and
  models,'' \emph{International Review of Financial Analysis}, vol.~33, pp.
  171--185, 2014.

\bibitem{gupta2020sentiment}
R.~Gupta and M.~Chen, ``Sentiment analysis for stock price prediction,'' in
  \emph{2020 IEEE conference on multimedia information processing and retrieval
  (MIPR)}.\hskip 1em plus 0.5em minus 0.4em\relax IEEE, 2020, pp. 213--218.

\bibitem{satapathy2020review}
R.~Satapathy, E.~Cambria, A.~Nanetti, and A.~Hussain, ``A review of shorthand
  systems: From brachygraphy to microtext and beyond,'' \emph{Cognitive
  Computation}, vol.~12, pp. 778--792, 2020.

\bibitem{duong2023wildfires}
C.~Duong, V.~Chithrra~Raghuram, A.~Lee, R.~Mao, G.~Mengaldo, and E.~Cambria,
  ``Neurosymbolic {AI} for mining public opinions about wildfires,''
  \emph{Cognitive Computation}, 2023.

\bibitem{mengaldo2023}
H.~Turb{\'e}, M.~Bjelogrlic, C.~Lovis, and G.~Mengaldo, ``Evaluation of
  post-hoc interpretability methods in time-series classification,''
  \emph{Nature Machine Intelligence}, vol.~5, no.~3, pp. 250--260, 2023.

\bibitem{xing2023guest}
F.~Xing, I.~Chaturvedi, E.~Cambria, A.~Hussain, and B.~Schuller, ``Guest
  editorial: Neurosymbolic {AI} for sentiment analysis,'' \emph{IEEE
  Transactions on Affective Computing}, vol.~14, no.~4, 2023.

\bibitem{cambria2022senticnet}
E.~Cambria, Q.~Liu, S.~Decherchi, F.~Xing, and K.~Kwok, ``Senticnet 7: A
  commonsense-based neurosymbolic ai framework for explainable sentiment
  analysis,'' in \emph{Proceedings of the Thirteenth Language Resources and
  Evaluation Conference}, 2022, pp. 3829--3839.

\end{thebibliography}

\end{document}